%% file: ms.tex
\documentclass{article}
\usepackage{microtype}
\usepackage{graphicx}
\usepackage{booktabs} 
\usepackage{float}
\usepackage{array}
\newcolumntype{L}[1]{>{\raggedright\arraybackslash}p{#1}}
\usepackage{amsmath,amssymb,amsfonts}
\usepackage{setspace}
\usepackage[table]{xcolor}
\usepackage{colortbl}
\usepackage{tabularx}  
\usepackage{calc} 
\usepackage{subcaption}
\usepackage{times}
\usepackage{algorithm}          
\usepackage[noend]{algpseudocode} 
\usepackage{epsfig}
\usepackage{threeparttable}
\usepackage{makecell}
\usepackage{multirow}
\usepackage{url}            
\usepackage{amsfonts}       
\usepackage{nicefrac}       
\usepackage{amsmath}
\usepackage{bbding}
\usepackage{float}
\usepackage{caption}
\usepackage{subcaption}
\usepackage{wrapfig}

\usepackage{textcomp}
\usepackage{xcolor}
\usepackage{listings}
\usepackage{adjustbox}
\usepackage{helvet}
\usepackage{booktabs}
\usepackage[table]{xcolor} 
\usepackage[none]{hyphenat}
\usepackage[numbers]{natbib}
\newcommand{\titleicon}{\raisebox{-0.28\height}{\includegraphics[width=1.3cm]{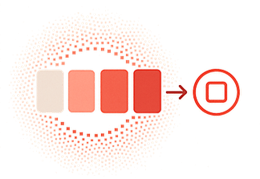}}}
\newcommand{\RunningTitleIcon}{\raisebox{-0.09\height}{\includegraphics[width=0.35cm]{Figures/icon.png}}}
\usepackage{hyperref}
\newcommand{\IconSep}{0.35em}      
\newcommand{\IconRowHt}{1.0em}     
\newcommand{\IconYDrop}{-0.45ex}   

\DeclareRobustCommand{\icon}[2][\IconSizeDefault]{%
  \raisebox{\IconYDrop}{\includegraphics[height=#1em]{#2}}%
}
\DeclareRobustCommand{\LongContextIcon}[1][\IconSizeDefault]{\icon[#1]{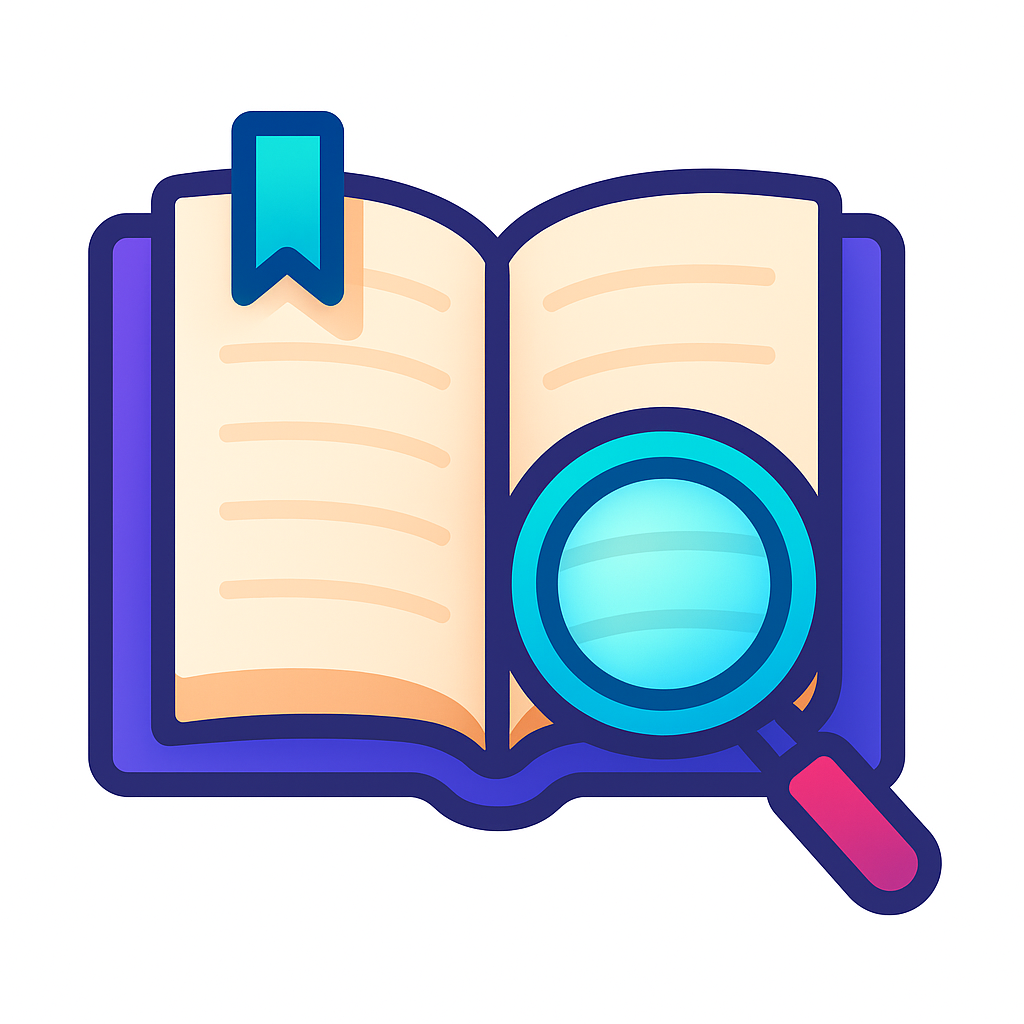}}
\DeclareRobustCommand{\LongGenIcon}[1][\IconSizeDefault]{\icon[#1]{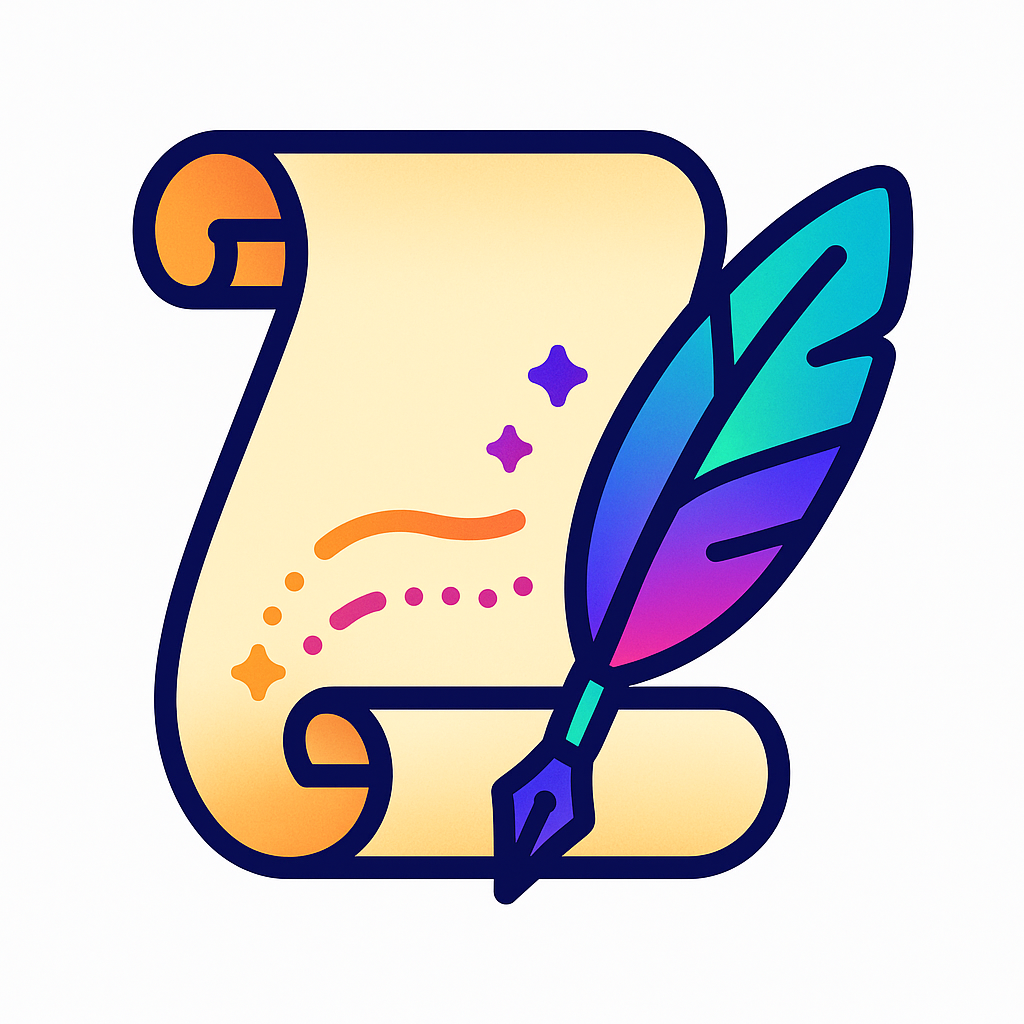}}

\newcommand{\HdrIcon}[3][\IconSizeDefault]{
  \rule{0pt}{\IconRowHt}
  \mbox{#2[#1]\hspace{\IconSep}#3}%
}

\definecolor{macpink}{HTML}{FCE4EC} 

\definecolor{secondcell}{gray}{0.97}
\newcommand{\kvdown}{\ensuremath{KV^\downarrow_\%}}

\newcommand{\E}{\mathbb{E}}
\newcommand{\R}{\mathbb{R}}
\newcommand{\norm}[1]{\left\lVert #1 \right\rVert}
\newcommand{\RoPE}{\textsc{RoPE}}

\newcommand{\GQA}{\textsc{GQA}}
\newcommand{\MQA}{\textsc{MQA}}
\newcommand{\MAC}{\textsc{MAC}\text{-Attention}}

\newcommand{\na}{\multicolumn{1}{c}{--}} 

\definecolor{RowGray}{gray}{0.98}
\rowcolors{4}{RowGray}{white} 

\usepackage[noend]{algpseudocode}  
\algrenewcommand\algorithmicrequire{\textbf{Input:}}
\algrenewcommand\algorithmicensure{\textbf{Output:}}

\usepackage[accepted]{mlsys2025}

\mlsystitlerunning{\RunningTitleIcon~MAC-Attention: a Match--Amend--Complete Scheme for Fast and Accurate Attention Computation}

\begin{document}

\twocolumn[
\mlsystitle{\titleicon~MAC-Attention: a Match--Amend--Complete Scheme for Fast and Accurate Attention Computation}



\mlsyssetsymbol{equal}{*}

\begin{mlsysauthorlist}
\mlsysauthor{Jinghan Yao}{osu,ms}
\mlsysauthor{Sam Ad\'{e} Jacobs}{ms}
\mlsysauthor{Walid Krichene}{ms}
\mlsysauthor{Masahiro Tanaka}{as}
\mlsysauthor{Dhabaleswar K Panda}{osu}
\end{mlsysauthorlist}

\mlsysaffiliation{osu}{The Ohio State University}
\mlsysaffiliation{ms}{Microsoft, WA, USA}
\mlsysaffiliation{as}{Anyscale, CA, USA}

\mlsyscorrespondingauthor{Jinghan Yao}{yjhmitweb@gmail.com}

\mlsyskeywords{Machine Learning, MLSys}

\vskip 0.3in
\input{Content/0-abstract}
]



\printAffiliationsAndNotice{}  

\input{Content/1-introduction}

\input{Content/2-background}
\input{Content/3-design}
\input{Content/3-5-SystemImplementation}

\input{Content/4-evaluation}

\input{Content/5-related_works}
\input{Content/6-discussion}
\input{Content/7-conclusion_and_future_work}

\input{Content/8-acknowledgements}
\bibliography{main}
\bibliographystyle{mlsys2025}

\appendix
\input{Content/9-appendix}


\end{document}

%% file: Content/0-abstract.tex
\begin{abstract}
Long-context decoding in LLMs is \emph{IO-bound}: each token re-reads an ever-growing KV cache. Prior accelerations cut bytes via \emph{compression} (lowering fidelity) or \emph{selection/eviction} (restricting what remains accessible), which can degrade delayed recall and long-form generation. We introduce \textbf{MAC-Attention}, a \emph{fidelity and access-preserving} alternative that accelerates decode by \emph{reusing prior attention computations} for \emph{semantically similar} recent queries. It starts with a \emph{match} stage that performs pre-RoPE L2 matching over a short local window; an \emph{amend} stage rectifies the reused attention by recomputing a small band near the match boundary; and a \emph{complete} stage fuses the rectified results with a fresh attention computed on the KV tail, via a numerically stable merge. On a match hit, the compute and bandwidth complexity is \emph{constant regardless of the context length}. The method is model-agnostic, and composes with IO-aware kernels, paged-KV managers, and MQA/GQA. Across LongBench~v2 (120K), RULER (120K), and LongGenBench (16K continuous generation), compared to latest FlashInfer library, \MAC\ reduces KV accesses by up to \textbf{99\%}, cuts token generation latency by over \textbf{60\%} at 128K, and achieves over \textbf{14.3$\times$} attention-phase speedups (up to \textbf{2.6$\times$} end-to-end), while maintaining full-attention quality. By reusing computation, \MAC\ delivers long-context inference that is both \emph{fast} and \emph{faithful}. Code is available \href{https://github.com/YJHMITWEB/MAC-Attention.git}{here}.
\end{abstract}


%% file: Content/1-introduction.tex
\section{Introduction}
Large Language Models (LLMs) now operate over tens to hundreds of thousands of tokens, enabling long‑document understanding, multi‑turn dialogue, codebase analysis, and scientific literature processing at unprecedented scales. Despite impressive kernel and runtime engineering, long‑context \emph{decoding} remains dominated by two factors: repeatedly streaming an ever‑growing Key–Value (KV) cache from high‑bandwidth memory (HBM) and reducing attention over long prefixes at every generated step. IO‑aware attention kernels~\cite{dao2022flashattention,dao2023flashattention2,ye2025flashinfer} and memory managers such as vLLM’s PagedAttention~\cite{kwon2023pagedattention} reduce waste and improve locality; nevertheless, the cost of re‑reading and processing large KV regions continues to be a primary latency bottleneck at long lengths.

\begin{figure}[t]
  \centering
  \includegraphics[width=0.47\textwidth]{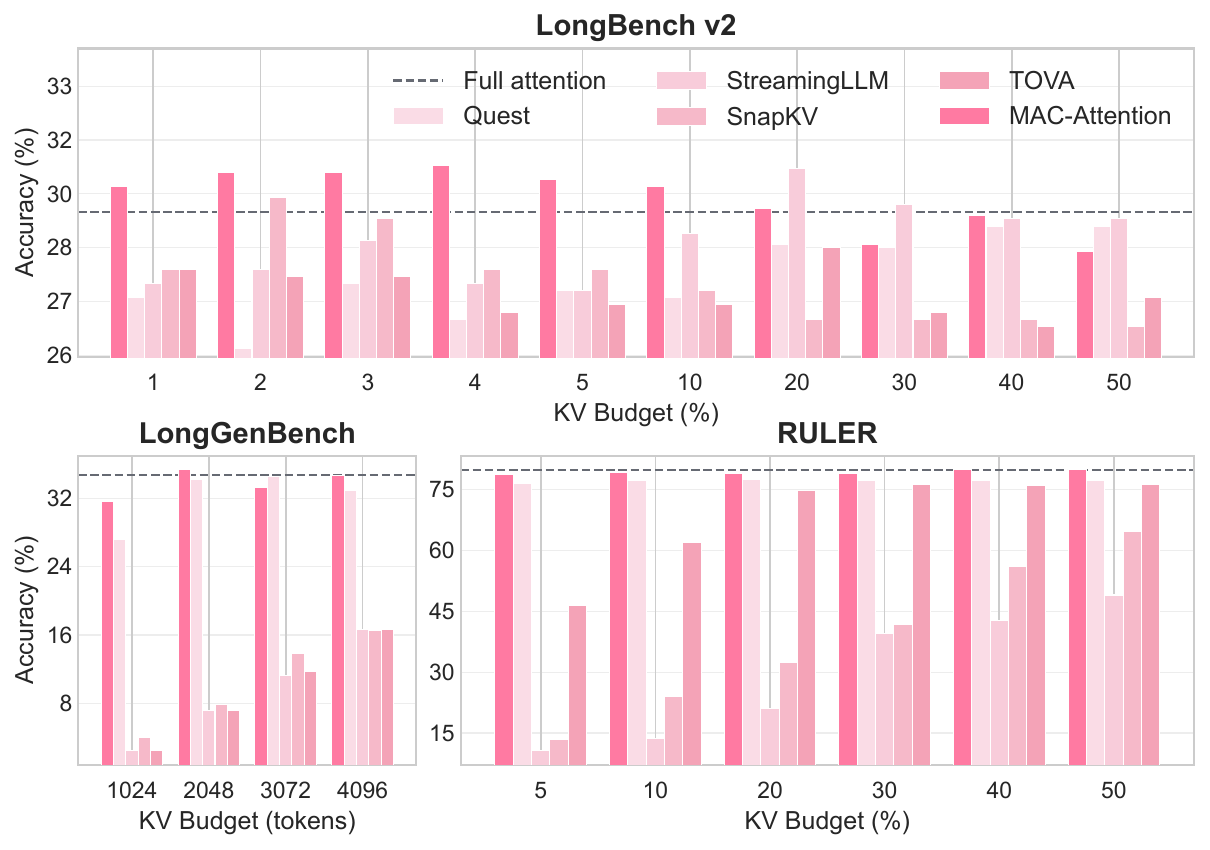}
  \vspace{-0.75em}
  \caption{\textbf{Accuracy vs.\ KV budget (the fraction of KV cache used in each decoding step) across long‑context benchmarks.}
  Top: \emph{LongBench v2} (up to 120K context)~\cite{bai2024longbenchv2}.
  Bottom left: \emph{LongGenBench} (up to 16K continuous generation)~\cite{wu2024longgenbench}.
  Bottom right: \emph{RULER} (120K context)~\cite{hsieh2024ruler}.
  \emph{MAC‑Attention} is highlighted (dark pink); \emph{Full attention} shown as a gray dashed line.}
  \label{fig:compare}
  \vspace{-.2in}
\end{figure}

Two families of approaches mitigate this IO bottleneck. The first compresses KV states (e.g., low‑rank projection or quantization) to reduce footprint and bandwidth~\cite{chang2024palu,zhang2024lorc}. The second selects or evicts tokens/pages (often adaptively) to shrink the effective context~\cite{xiao2024streamingllm,zhang2023h2o,liu2023scissorhands,li2024snapkv,ge2023fastgen,feng2024adakv,tang2024quest}. Both are effective at lowering bytes moved, but they introduce approximation errors either by constraining the range of tokens the model can attend to (selection/eviction), or by reducing fidelity (compression).
As a result, performance on tasks with delayed recall, cross‑document linking, long generation/reasoning, or distributed evidence can degrade~\cite{hsieh2024ruler}, as shown in Fig.~\ref{fig:compare}.

This paper introduces \textbf{MAC‑Attention (MAC)}—a training‑free, model‑agnostic mechanism that accelerates long‑context decoding by \emph{reusing prior attention computation} for \emph{semantically similar} recent queries while preserving access to the full sequence, and achieving high fidelity. 
MAC maintains two short‑horizon ring buffers per request: a pre‑positional (pre‑RoPE) \emph{query ring} and a corresponding \emph{rectified attention‑summary ring}. When a new query matches one in the window, MAC reuses the cached attention-summary of the prefix, computes the contribution of the tail, then \emph{merges} them via an \emph{associative, numerically stable log‑domain merge}~\cite{milakov2018online,dao2022flashattention}. Whenever there is a match, the computation (and memory bandwidth) are \emph{constant} in the sequence length. When there is no match, it falls back to full attention computation (linear).

\textbf{How MAC Attention maintains high fidelity.} 
There are two algorithmic improvements that are crucial to maintaining a high fidelity (i.e., a low approximation error of the attention output). The first improvement is how query matching is done. Most modern LLMs use positional encodings that express relative position via rotations or distance‑dependent biases, e.g., RoPE and ALiBi~\cite{su2021roformer,press2021alibi}. While it may be simpler to match in post‑position (post‑RoPE) space, we find that due to the effect of position encodings, distances between otherwise similar queries can become large in post-RoPE space, which significantly reduces close matches. Instead, MAC \emph{matches in pre‑RoPE space}, which increases the matching rate and results in matches that are semantically close (regardless of position). The second improvement is the observation that tokens that are near the match position account for a significant fraction of the approximation error introduced by reusing the cached output. This is because the softmax probability mass typically concentrates near the decoding cursor~\cite{su2021roformer,press2021alibi}. Thus MAC \emph{rectifies the cached output by recomputing a short band} (of width \(r\)) around the match, significantly reducing approximation errors, while only paying a constant \(O(r)\) cost per‑head.

\textbf{Relation to other reuse paradigms.} Beyond compression/selection, prior work reuses structure across requests (e.g., prefix caching or tree‑structured decoding)~\cite{gim2024promptcache,yao2025deft} or alternates full and partial steps by recycling top‑\(k\) attention from previous steps~\cite{xu2024recycled}. MAC is orthogonal: it performs \emph{in‑stream semantic reuse with local rectification}, independent of literal prefix overlap or alternating step schedules, and preserves access to the full context.

\textbf{Results at a glance.} Across \emph{LongBench v2} (120K)~\cite{bai2024longbenchv2}, \emph{RULER} (120K)~\cite{hsieh2024ruler}, and \emph{LongGenBench} (16K continuous generation)~\cite{wu2024longgenbench}, MAC \emph{reduces KV accesses by up to 99\%}, cuts \emph{per‑token latency by over 60\% at 128K}, achieves \(\ge\!14.3\times\) attention‑phase speedups (up to \(\sim\!46\times\) under 256K context settings), and delivers \emph{up to \(2.6\times\)} end‑to‑end generation speedups on LLaMA—\emph{while maintaining full‑attention quality}.

Contributions are as follows:
\begin{itemize}
  \item We propose \emph{Match–Amend–Complete (MAC) Attention}, a fidelity and access‑preserving reuse scheme that \emph{matches} pre‑RoPE queries using an L2, dimension‑aware threshold, \emph{amends} a short high‑mass band near the reuse boundary, and \emph{completes} with an associative, numerically stable \emph{log‑domain merge}.
  \item MAC is \emph{training‑free}, \emph{model‑agnostic}, and \emph{drop‑in} for high‑performance inference stacks, composing with IO‑aware attention~\cite{dao2022flashattention,dao2023flashattention2,ye2025flashinfer}, optimized decode paths~\cite{hong2024flashdecodingpp}, paged KV~\cite{kwon2023pagedattention}, and MQA/GQA~\cite{shazeer2019mqa,ainslie2023gqa}.
  \item Across LongBench v2, RULER, and LongGenBench~\cite{bai2024longbenchv2,hsieh2024ruler,wu2024longgenbench}, MAC \emph{reduces KV accesses by up to 99\%}, achieves \(\ge\!14.3\times\) (up to \(\sim\!46\times\)) attention‑phase speedups and \emph{up to \(2.6\times\)} end‑to‑end speedups on LLaMA, while maintaining full‑attention quality.
\end{itemize}

%% file: Content/2-background.tex
\section{Background}
\label{sec:background}

\paragraph{Long-context decode}
Even with I/O-aware kernels, the dominant cost at long context decoding remains memory traffic: streaming large KV regions from high-bandwidth memory (HBM)~\cite{dao2022flashattention,hong2024flashdecodingpp,dao2023flashdecoding}. To lower bytes moved, existing work either (i) \emph{compresses} KV states (e.g., by projection or quantization), or (ii) \emph{selects/evicts} tokens or pages to shrink the effective context~\cite{hooper2024kvquant,liu2024minicache,ge2024adaptive,zhang2023h2o,liu2023scissorhands,xiao2024streamingllm}. Both strategies can substantially cut I/O, but they also change either how information is represented (compression) or which tokens remain accessible (selection/eviction).

\paragraph{Exploiting temporal redundancy} This paper explores a third path, rooted in the observation that computation within a single decoding stream often exhibits high \textbf{temporal redundancy}~\cite{chen2024arkvale,liu2023scissorhands,xu2024recycled}. In many generative tasks, such as multi-turn dialogue, code generation, or long reasoning, the model's internal state-represented by the query vector at each step-is not drawn from a uniform distribution. Instead, queries often perform repetitive information-retrieving patterns. This redundancy presents a clear opportunity for reuse. The attention distribution of two similar queries induced over the shared prefix is likely to be highly correlated. Rather than re-streaming attention over the shared prefix, reusing it becomes a clear and intuitive path.



\paragraph{Notation}
Let $Q_t, K_t, V_t$ denote, respectively, the query, key and value encodings for token position $t$ (our discussion applies to any attention block, so we omit indexing of layers or heads for notational simplicity). We also denote by $\tilde Q_t$ the query encodings prior to applying any positional encodings (in short, the per-RoPE query). We will denote by $m$ the position of the current token, for which we seek to compute the attention output. Define the logits vector $\ell_t^{(m)} = \tfrac{1}{\sqrt d}\, Q_m^\top K_t$. The attention output is then defined as follows: $o_m = S^{(m)}/Z^{(m)}$ where $S^{(m)}$ is the weighted sum $S^{(m)} = \sum_{t \leq m} e^{\ell_t^{(m)}} V_t$, and $Z^{(m)}$ is the normalizing constant $Z^{(m)} = \sum_{t \leq m} e^{\ell_t^{(m)}}$.

%% file: Content/3-design.tex
\section{MAC-Attention: Design}
\label{sec:design}

\subsection{Design overview}
Our design starts with the following simple observation: that the attention computation can be decomposed, at any position $p$, into a prefix and suffix computations. For a set $\mathcal I$, Define the \emph{attention summary}%
{\small\[
AS^{(m)}_{\mathcal I} \;=\; \Big(S^{(m)}_{\mathcal I},\; Z^{(m)}_{\mathcal I}\Big)
\;=\;
\Big(\textstyle\sum_{t\in\mathcal I} e^{\ell_t^{(m)}} v_t,\;\sum_{t\in\mathcal I} e^{\ell_t^{(m)}}\Big).
\]}%
The exact output is $o_m = S_{1:m}^{(m)}/Z_{1:m}^{(m)}$. Summaries over disjoint sets merge by addition. For any position $p$, attention computation can be decomposed into $o_m = \frac{S_{1:p}^{(m)}+S_{p+1:m}^{(m)}}{Z_{1:p}^{(m)} + Z_{p+1:m}^{(m)}}$. This decomposition is numerically stable and associative, and the special case $p = m-1$ is used in popular methods such as Flash Attention~\citep{milakov2018online,dao2022flashattention}.

The starting point of the design is to attempt to \emph{cache} the results of similar past computations and use them in the decomposition above. Specifically, we will attempt to reuse the \emph{prefixes} $S_{1:p}^{(m)}$ and $Z_{1:p}^{(m)}$, and approximate them by $S_{1:p}^{(p)}, Z_{1:p}^{(p)}$. These quantities are similar but not identical: the first is computed based on the logits $\ell_t^{(m)} = K_t^\top Q_m, t\leq p$ while the second uses $\ell_t^{(p)} = K_t^\top Q_p, t\leq p$ (they share the keys but differ in the queries). However, notice that if the current query $Q_m$ is close to $Q_p$, their logits are also close, and $S_{1:p}^{(p)}$ can be a good approximation of the true~$S_{1:p}^{(m)}$ (and similarly for $Z$). This forms the basis of {our method.}

The next crucial observation is that in order to further reduce approximation errors, we can discard and \emph{recompute} a very small part of the prefix (the interval near the match position $[p-r, p]$, which in practice accounts for most of approximation error for reasons we shall discuss). More precisely, we use the \emph{truncated} prefix $S_{1:p-r}^{(p)}$ as an approximation of $S_{1:p-r}^{(m)}$, and freshly compute $S_{p-r+1:m}^{(m)}$. This only slightly increases the KV access (by a constant $r$) while significantly reducing approximation errors.

\begin{figure*}[thbp]
    \centering
    \includegraphics[width=0.98\textwidth,page=1]{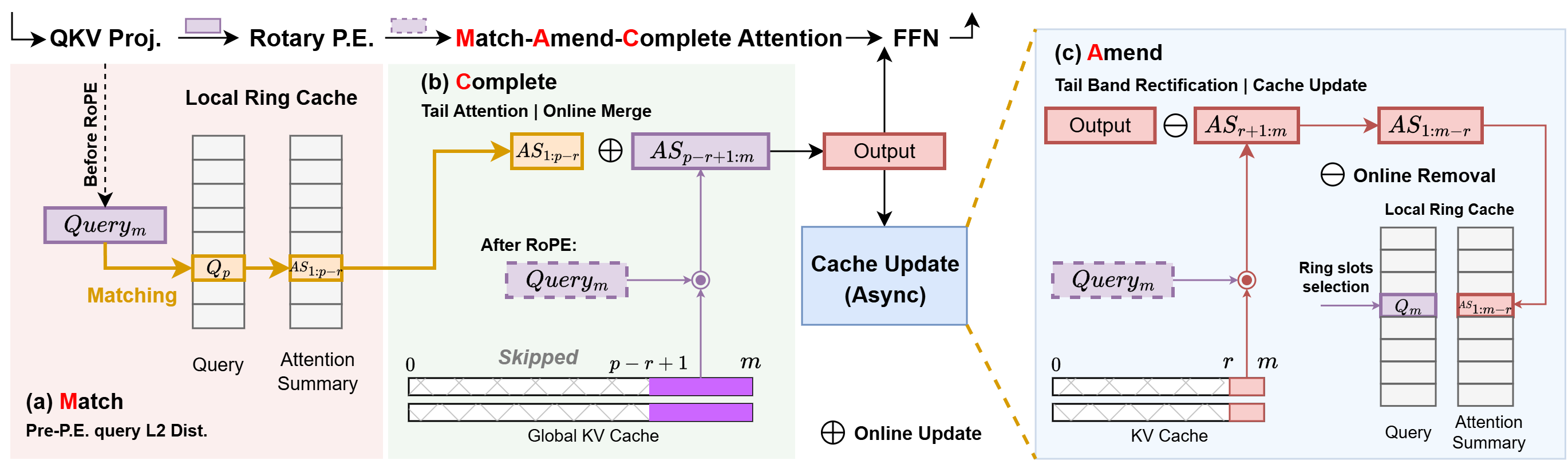}
    \caption{\textbf{MAC‑Attention: Match–Amend–Complete with async cache update.}
    \textbf{(a) Match:} at position $m$, compare the \emph{pre‑RoPE} query $\tilde Q_m$ against a small ring of recent queries; if the nearest match $p$ passes an L2 threshold, fetch its attention summary $\mathrm{AS}_{1:p-r}$ (otherwise run full attention).
    \textbf{(b) Complete (critical path):} with the \emph{post‑RoPE} query $q_m$, compute attention only on the band+tail $[\,p{-}r{+}1,\,m\,]$ and log‑domain merge $\mathrm{AS}_{1:p-r} \oplus \mathrm{AS}_{p-r+1:m}$; the KV $[1,p{-}r]$ is not accessed.
    \textbf{(c) Amend (async):} compute a rectification term $\mathrm{AS}_{r+1:m}$ and update the cache via online subtraction to obtain $\mathrm{AS}_{m-r}$; insert $\tilde q_m, \mathrm{AS}_{m-r}$ into the rings.
    Symbols: $p$—match position, $r$—band width; $\oplus$—log‑domain merge, $\ominus$—log‑domain removal. Shaded regions denote KV segments not re‑read under reuse.}
    \label{fig:workflow}
\end{figure*}

Figure~\ref{fig:workflow} summarizes the method: \MAC\ keeps two additional caches: a past query cache (for $Q_t$) and a past attention result cache (for the truncated/rectified attention summaries $S, Z$). In decoding step $m$, we first perform a past-query \emph{match}; if a matching query $Q_p$ is found ($p<m$), we reuse the corresponding summaries, which carry context information up to $p-r$. Then, we \emph{complete} the result by computing the suffixes $S_{p-r+1:m}^{(m)}, Z_{p-r+1:m}^{(m)}$ and online-merge to obtain the final output.

In the remainder of this section, we discuss design choices and additional details for each step.


\subsection{Match: where and how to compare}
\label{subsec:match}
In this part, we discuss details of the match strategy.

\subsubsection{Pre vs.\ post position match}
A first question is at what stage should the query match be performed. Indeed, in most recent models, positional encodings are applied to the query after QKV projection, and a natural question is whether the match should occur before or after the positional encodings. Let $\tilde Q_t\in\R^{d}$ be the \emph{pre‑positional} query at position $t$, and let $Q_t = R(t)\tilde Q_t$ be the \RoPE‑transformed query with $R(\cdot)$ orthogonal~\citep{su2021roformer}.

From the perspective of the operation order (\emph{QKV projection} $\rightarrow$ \emph{\RoPE} $\rightarrow$ \emph{attention}), post‑\RoPE\ matching seems more natural in principle because attention consumes the post‑position vectors. Writing the logits at position $m$ as
\(
\ell_t^{(m)} = \tfrac{1}{\sqrt d}\, Q_m^\top K_t
\)
with $Q_m=R(m)\tilde Q_m$ and $K_t=R(t)\tilde K_t$~\citep{vaswani2017attention}, Cauchy–Schwarz yields, for any key $K_t$, $
\big|\ell_t^{(m)} - \ell_t^{(p)}\big|
= \frac{\big|(Q_m - Q_p)^\top K_t\big|}{\sqrt d}
\le \frac{\|Q_m - Q_p\|\,\|K_t\|}{\sqrt d}$.
Thus a smaller post‑\RoPE\ distance directly bounds the \emph{worst‑case} logit error across all keys. 

In practice, however, due to the relative rotation $R(m{-}p)$, as $|m{-}p|$ grows, the post-\RoPE\ distance between similar queries tends to increase, \emph{reducing match success}. Indeed, matching in post‑\RoPE\ space injects a \emph{relative} phase $R(m{-}p)$ into the comparison. For $m>p$,
\begin{equation}
\label{eq:post-rot}
\norm{Q_m - Q_p}^2 \;=\; \norm{\tilde Q_m - R(m-p)\tilde Q_p}^2.
\end{equation}

Even if $\tilde Q_m \!\approx\! \tilde Q_p$, the cross‑term $-2\,\tilde Q_m^\top R(m{-}p)\tilde Q_p$ diminishes as $\Delta = |m{-}p|$ grows because \RoPE\ is block‑diagonal in $2{\times}2$ rotations:%
{\small\begin{multline}
R(\Delta)=\mathrm{diag}\big(
  R_{\omega_1}(\Delta),\ldots,R_{\omega_{d/2}}(\Delta)
\big),\\
x^\top R(\Delta) x
  \;=\; \sum_{j=1}^{d/2} \cos(\omega_j \Delta)\,\|x_{(j)}\|^2,
\end{multline}}%
with $x_{(j)}\in\R^2$ the $j$‑th frequency pair. When energy is distributed across frequencies, the weighted average
\[
\mathrm{avg\_cos}(\Delta) \;\equiv\; \frac{\sum_j \cos(\omega_j\Delta)\,\|x_{(j)}\|^2}{\|x\|^2}
\]
becomes small for moderate $\Delta$ due to destructive phase mixing, and
$\|x-R(\Delta)x\|^2 = 2\|x\|^2\big(1-\mathrm{avg\_cos}(\Delta)\big)$ approaches $2\|x\|^2$. Consequently, $\|Q_m{-}Q_p\|$ tendsd to increase with the gap $|m{-}p|$, potentially reducing matches even for semantically similar pairs.

Our design, therefore, matches \emph{pre‑\RoPE} to measure semantic similarity without positional phase (Fig.~\ref{fig:workflow}\,a), then performs a short, local \emph{rectification} near the reuse boundary to recompute the logits that carry most softmax mass under the true $Q_m$ (see Section~\ref{subsec:amend}). This combination recovers the practical fidelity one would expect from post‑\RoPE\ matching while maintaining robust match frequency. If no match is found, fall back to full attention; outputs are identical to baseline.

\subsubsection{L2 vs. cosine match}
A second design choice is the similarity measure. Choosing the right measure is important to reduce approximation errors. Two natural choices are cosine and dot-product. 
Cosine similarity captures only \emph{direction}; two colinear queries with different norms have the same cosine yet induce different logit scales and therefore different softmax sharpness. In contrast, the Euclidean distance controls both angle and magnitude. This matters for fidelity as shown above, hence minimizing $\|Q_m - Q_p\|$ uniformly bounds the worst-case logit approximation error.

For isotropic pre-\RoPE\ queries, a simple Gaussian model \( \tilde Q_m-\tilde Q_p\sim\mathcal N(0,2I_d) \) implies \( \|\tilde Q_m-\tilde Q_p\|^2 \stackrel{d}{=} 2\,\chi^2_d \), so \( \|\tilde Q_m-\tilde Q_p\| \) concentrates near \( \sqrt{2d} \). We therefore accept a match when the L2 distance falls \emph{proportionally} below this baseline:%
{\small \begin{equation}
\label{eq:accept}
\|\tilde Q_m-\tilde Q_p\| \;<\; \sqrt{2d}\,(1-\tau), \qquad \tau\in[0,1),
\end{equation}}%
The parameter \(\tau\) defines a threshold relative to random coincidence, and is set to a fixed value across heads.

\subsubsection{Local window and search cost}
The candidate set is restricted to a short, recent window of size $K$ per (grouped) head.
There are several reasons for this: First, reuse is most valuable when the matched prefix is long (i.e. when the match position is near the current token). Second, this allows us to cache only a constant number of attention results, and match against a constant number of past queries (making the memory and compute cost of matching \emph{constant, rather than linear in the sequence length}). Last, similar queries are more likely to occur within a short horizon, so restricting to the last $K$ tokens does not significantly impact the match rate.

The matcher is a single streaming pass over the window: for each candidate $\tilde Q_p$ it accumulates two inner products to form the squared distance, $
\|\tilde Q_m-\tilde Q_p\|^2
=\|\tilde Q_m\|^2 + \|\tilde Q_p\|^2 - 2\,\tilde Q_m^\top \tilde Q_p$,
performing bf16 reads with fp32 accumulation. Arithmetic intensity is low, so the kernel is memory‑bound but lightweight; tiling over the ring and a warp‑only reducer ensures sufficient waves to hide latency (see \S\ref{sec:system-design}).

\subsection{Complete: attention summaries and online merge}
\label{subsec:complete}

As shown in Fig~\ref{fig:workflow}\,b, suppose the matcher selects index $p<m$ and we have cached $AS^{(p)}_{1:p-r}$. If we could replace $AS^{(m)}_{1:p-r}$ by $AS^{(p)}_{1:p-r}$ without loss, then
\begin{equation}
o_m \;=\; \frac{S^{(m)}_{1:p-r} + S^{(m)}_{p-r+1:m}}{Z^{(m)}_{1:p-r} + Z^{(m)}_{p-r+1:m}} \;\approx\; \frac{S^{(p)}_{1:p-r} + S^{(m)}_{p-r+1:m}}{Z^{(p)}_{1:p-r} + Z^{(m)}_{p-r+1:m}}.
\label{eq:merge}
\end{equation}
Equation~\eqref{eq:merge} is the essence of \emph{complete}: once a prefix is summarized, fusing it with a freshly computed tail is constant-time in the size of the tail.

\subsection{Amend: why rectification is necessary, and how to do it}
\label{subsec:amend}

\subsubsection{Where naive reuse errs}
To understand why rectification is important, consider the naive reuse strategy, that directly replaces the full prefix $AS^{(m)}_{1:p}$ with $AS^{(p)}_{1:p}$.
Let $\delta_t\!\equiv\!\ell_t^{(m)}-\ell_t^{(p)}$ for $t\le p$. Then
$S_{1:p}^{(m)}=\sum_{t\le p} e^{\ell_t^{(p)}+\delta_t}v_t$ and
$Z_{1:p}^{(m)}=\sum_{t\le p} e^{\ell_t^{(p)}+\delta_t}$. Using the (full) cached prefix corresponds to setting $\delta_t=0$.
A first-order expansion of softmax gives
\begin{equation}
\Delta \alpha_t \;\approx\; \alpha_t^{(p)}\Big(\delta_t - \E_{u\sim\alpha^{(p)}}[\delta_u]\Big),
\label{eq:jac}
\end{equation}
so the approximation error at position $k$ increases with $\alpha^{(p)}_k$. In practice, softmax weights are typically larger for \emph{recent} tokens near $p$ due to semantic similarity and recency bias from position encodings~\citep{su2021roformer,press2021alibi}. Thus, the approximation error is typically dominated by local tokens.


\subsubsection{Rectification as removing a short high-mass band}
Let $\mathcal R=\{p-r+1,\dots,p\}$ be a short band capturing most prefix mass, with $\rho=\sum_{t\in\mathcal R}\alpha^{(p)}_t$. Define the \emph{rectified} cached prefix by \emph{removing} this band at position $p$:
$AS^{(p)}_{1:p-r}$.

At position $m$, we recompute only $[p-r+1, m]$ \emph{under the new query}, and merge with the cached summary:
\begin{align}
S^{(m)}_{p-r+1:m} &= \sum_{t=p-r+1}^{m} e^{\ell_t^{(m)}} v_t, &
Z^{(m)}_{p-r+1:m} &= \sum_{t=p-r+1}^{m} e^{\ell_t^{(m)}}.
\label{eq:rect-recomp}
\end{align}
Then the final output is
\begin{equation}
o_m \;=\; \frac{S^{(p)}_{1:p-r} + S^{(m)}_{p-r+1:m}}{Z^{(p)}_{1:p-r} + Z^{(m)}_{p-r+1:m}}.
\label{eq:rect-merge}
\end{equation}
Removing and recomputing only the short high-mass band significantly reduces approximation errors, while keeping a constant $O(r)$ cost per head (independent of context length).

\subsubsection{Error decays with residual mass outside the band}
Using a Lipschitz bound for exponentials and \eqref{eq:jac}, one can show that the prefix error after rectification scales as
\begin{equation}
\big\|o^{\text{reuse}}_{1:p}-o^{(m)}_{1:p}\big\|
\;\lesssim\; (e^{\Delta}-1)\,(1-\rho)\cdot
\E_{t\sim\alpha^{(p)}_{\bar{\mathcal R}}}\!\big[\|v_t\|\big],
\label{eq:mass-bound}
\end{equation}
so choosing $r$ to capture most mass ($\rho\uparrow$) drives the error down quickly. In practice, a fixed small $r$ per head works well across lengths, as shown in Fig.~\ref{fig:workflow}\,c, we \emph{cache rectified summaries} at creation time (position $p$), making reuse a constant-time merge at position $m$ via \eqref{eq:rect-merge}.

\subsection{Putting it together: match + amend + complete}
\label{subsec:mac-pipeline}
Per query, per head, and per layer:
\begin{enumerate}
  \item \textbf{Match.} Search the recent window of $K$ pre-\RoPE\ queries using squared L2 and accept if \eqref{eq:accept} holds.
  \item \textbf{Amend.} Load the \emph{rectified} prefix summary $(S^{(p)}_{1:p-r},Z^{(p)}_{1:p-r})$ and recompute only the band $[p{-}r{+}1,p]$ and tail $(p, m]$ under the current query.
  \item \textbf{Complete.} Merge via the online identity \eqref{eq:rect-merge}. 
\end{enumerate}
We cache (i) the last $K$ pre-\RoPE\ queries and (ii) the corresponding $K$ \emph{rectified} prefix summary; Alone with the existing KV cache. In practice, $K\leq1024$, introducing a negligible memory overhead in long context scenarios.

\subsection{Economics: when does MAC reduce memory bandwidth?}
\label{subsec:economics}
Let $B_{\mathrm{KV}}$ be bytes read per cached token in attention and $B_q$ bytes read per candidate query during matching. If the match is at position $p$ with rectification width $r$ and we search $K$ candidates, a coarse break-even condition (in terms of memory traffic) is
\begin{equation}
\underbrace{p\,B_{\mathrm{KV}}}_{\text{prefix avoided}}
\;\gtrsim\;
\underbrace{K\,B_q}_{\text{match cost}}
\;+\;
\underbrace{r\,B_{\mathrm{KV}}}_{\text{band+tail recompute}}
\label{eq:roi}
\end{equation}
In practice, we will keep $K\leq1024$ and $r\leq512$ so that \MAC\ can skip a substantial amount of KV access while also maintain a full attention fidelity. We provide a thorough analysis in \S\ref{app:band_window_heatmap}.

\subsection{Practical notes and compatibility}
\textbf{Short-horizon rings.} We maintain per-request size $K$ ring buffers for queries and rectified summaries. The ring capacity bounds memory and ensures $O(1)$ insertion. 

\textbf{KV sharing.} \MQA/\GQA\ ~\cite{shazeer2019mqa,ainslie2023gqa} reduce the number of KV heads. \MAC\ matches and reuses at the query granularity as attention is performed per-query head, thus the matching kernel is more memory-bound then the attention kernel (See Figure~\ref{fig:kernel_breakdown}(a)).

%% file: Content/3-5-SystemImplementation.tex
\section{System design}
\label{sec:system-design}
\MAC\ targets that decoding IO-bound challenge directly by (i) matching to reuse a long prefix, (ii) amending only a short high‑mass band, and (iii) completing with an online merge. The system must therefore (1) keep the matching and amendment overhead below the saved KV traffic, (2) preserve numerical precision during multiple merges, and (3) integrate cleanly with IO‑aware attention and paged KV managers without perturbing batching or memory layout.

\subsection{State and data layout}

We maintain two short-horizon, per-request ring buffers, each with a capacity of $K$ tokens. The \emph{query ring} stores pre-\RoPE\ queries from the most recent $K$ tokens for each request. The \emph{attention-summary ring} preserves the rectified prefix summaries $(S^{\mathrm{rect}}_p, Z^{\mathrm{rect}}_p)$. Both rings are indexed by the request’s global length in modulo $K$ order, so the recent window is contiguous and insertion is $O(1)$. The scheduler maps multiple query heads to the corresponding KV head without changing the underlying layout.

\begin{figure}[t]
  \centering
  \includegraphics[width=\linewidth]{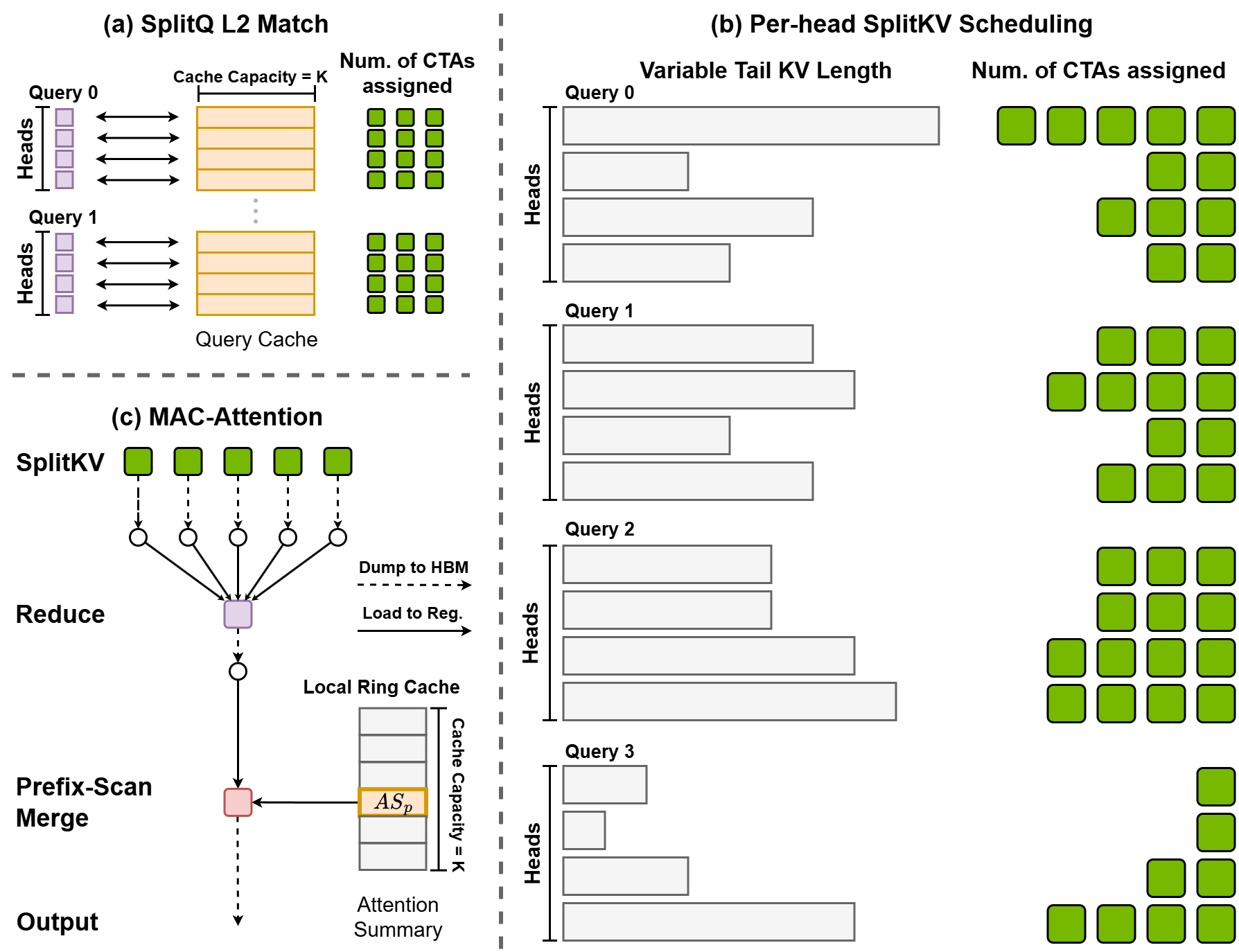}
  \caption{\textbf{Decode micro‑pipeline for \MAC.} \emph{(a):} per‑request rings L2 match with SplitQ design; \emph{(b):} per‑head work spans differ because the reuse point $p$ and band size $r$ vary, the schedule assigns more CTAs to longer band+tail spans as shown in green blocks; many heads do a cheap merge when reuse is strong; \emph{(c):} Overview of \MAC\ workflow.}
  \label{fig:kernel-design}
  \vspace{-1.0em}
\end{figure}

\begin{table*}[t]
  \centering
  \caption{\textbf{Main accuracy results across models and benchmarks.} Accuracy (\%) on LongBench v2 (up to 120K context), LongGenBench (up to 16K continuous generation), and RULER (120k context). Rows compare \emph{Full attention}, \emph{Quest} variants (config / shows the budget settings for each benchmark; \emph{Quest} only supports LLaMA models.), and our \emph{MAC-Attention}. The \(\kvdown\) columns report average KV-budget reduction relative to full attention (higher is more savings); \texttt{-} denotes not applicable. LongGenBench is generation-heavy; most models cannot complete the required task; we therefore report \emph{Finish\%} (higher is better), and finish-weighted accuracy.}

  \label{tab:acc_8b}
    \begin{threeparttable}
  \centering
  \small
  \begin{tabular}{@{} ll c c c c c c c c c c @{}}
    \toprule
    & & \multicolumn{5}{c}{\HdrIcon{\LongContextIcon}{LongBench v2}} & \multicolumn{3}{c}{\HdrIcon{\LongGenIcon}{LongGenBench}} & \multicolumn{2}{c}{\HdrIcon{\LongContextIcon}{RULER}} \\
    \cmidrule(lr){3-7}\cmidrule(lr){8-10}\cmidrule(lr){11-12}
    Model & Config. &
    {Overall} & {Short} & {Medium} & {Long} & {$\kvdown$} &
    {Finish~$\%$} & {Acc.} & {$\kvdown$} &
    {Acc.} & {$\kvdown$} \\
    \midrule
    \textbf{LLaMA-3.1-8B} & Full attention   & 29.0 & 35.0 & 26.0 & 25.0 &0 &   93.1 & 34.7 & 0 & 79.8 & 0 \\
     \cite{meta2024llama3} & Quest-8K/4K/8K  & 28.0 & 33.9 & 24.2 & 25.9 & 85 & 92.8 & 32.9 & 35 & 77 & 93 \\
      & Quest-4K/3K/4K  & 27.4 & 33.3 & 24.7 & 23.1 & 92 & 91.6 & 34.8 & 48 & 75.5 & 97 \\
      & Quest-2K  & 27.8 & 32.2 & 26.0 & 24.1 & 96 & 89.9 & 33.8 & 63 & 73.4 & 98 \\
      & Quest-1K  & 26.2 & 33.3 & 22.3 & 22.2 & 98 & 83.6 & 27.2 & 80 & 72.4 & 99 \\
      \rowcolor{macpink}
      & \textbf{MAC-Attention}  & 30.2 & 34.4 & 27.4 & 28.7 & 99 & 90.0 & 38.2 & 80 & 78.8 & 95 \\
    \midrule

    \textbf{Phi-4-Mini} & Full attention   & 28.0 & 32.8 & 23.3 & 29.6 & 0 & \na & \na & \na & 74.4 & 0 \\
    \rowcolor{macpink}
     \cite{microsoft2025phi4mini} & \textbf{MAC-Attention}  & 29.8 & 33.9 & 25.1 & 32.4 & 91 & \na & \na & \na & 73.1 & 77 \\
    \midrule
    
    \textbf{GLM-4-32B} & Full attention   & \na & \na & \na & \na & \na & 69.8 & 29.7 & 0 & \na & \na \\
    \rowcolor{macpink}
      \cite{zhipu2025glm432b}&  \textbf{MAC-Attention}  & \na & \na & \na & \na & \na & 66.1 & 30.1 & 70 & \na & \na \\
    \midrule

    \textbf{LLaMA-3.1-70B} & Full attention   & 31.4 & 41.7 & 26.5 & 24.1 & 0 & 98.1 & 41.9 & 0 & 80.1 & 0 \\
      \cite{meta2024llama3}& Quest-8K/4K/8K  & 31.2 & 40.6 & 27.4 & 23.1 & 85 & 97.5 & 45.8 & 29 & 79.8 & 93 \\
      & Quest-4K/3K/4K  & 31.8 & 40.6 & 27.9 & 25.0 & 92 & 98.6 & 45.2 & 40 & 79.8 & 97 \\
      & Quest-2K  & 31.2 & 39.4 & 27.9 & 22.2 & 96 & 98.3 & 46.2 & 54 & 79.5 & 98 \\
      & Quest-1K  & 30.8 & 39.4 & 27.9 & 22.2 & 98 & 93.8 & 41.9 & 72 & 78.3 & 99 \\
      \rowcolor{macpink}
      & \textbf{MAC-Attention}  & 32.2 & 40.6 & 28.8 & 25.0 & 99 & 98.6 & 43.6 & 75 & 78.0 & 99 \\
    \bottomrule
  \end{tabular}

  \begin{tablenotes}[flushleft]
    \scriptsize
    \item[] GLM-4-32B has 32K context limit, Phi-4-Mini baseline generates poor-quality responses on LongGenBench, hence their results are excluded from the table.
  \end{tablenotes}
  \end{threeparttable}
\end{table*}

\subsection{Per‑step micro‑pipeline}
In each decode step, \MAC\ executes three kernels with minimal synchronization:

\medskip
\noindent\textbf{K1. Match.} A tiled, L2 nearest‑neighbor scan compares each active query against the local window in its request’s query ring (pre‑\RoPE) as shown in Fig.~\ref{fig:kernel-design}\,(a). We emit per‑tile minima and run a warp‑only final reducer to choose \emph{at most} one reuse point $p$ per (request, head), then apply the dimension‑aware threshold from Eq.~\eqref{eq:accept}. Implementation choices are geared to IO efficiency. The reducer is shuffle‑based to eliminate barrier stalls. Compared to standard attention kernels with MQA/GQA, the match kernel is intrinsically more memory-bound as it needs to stream all query heads.

\medskip
\noindent\textbf{K2. Amend \& complete.} \emph{(1) Workload shaping and load balance:} Because each head may reuse a different prefix, the band+tail span varies per head. We therefore flatten the query–head axis and allocate CTAs proportionally to workloads (Fig.~\ref{fig:kernel-design}\,b), giving longer spans more thread blocks. \emph{(2) Attention computation:} Given a hit, compute attention only over the band+tail under the current query with the effective KV. \emph{(3) Merge:} Load the rectified prefix summary from the attention‑summary ring and merge in place.

\medskip
\noindent\textbf{K3. Rectify–append (ultra-fast, off‑critical‑path).} To prepare future reuse, we \emph{construct the next rectified summary} for the just‑produced token and append it, together with the pre‑\RoPE\ query, to the rings. K3 writes three rows: the query, the rectified attention row, and the ln‑LSE scalar. Crucially, K3 \emph{runs on an auxiliary stream} with a single event dependency—the result is not needed until the next step visits the same layer. Crucially, as we implemented K3 with fully in-place operations, it also can run within the main stream as its latency is negligible.

The overall kernel execution order and HBM traffic within the attention operation is shown in Fig.~\ref{fig:kernel-design}\,(c). \MAC\ composes with IO-aware kernels~\citep{dao2022flashattention,dao2023flashattention2,ye2025flashinfer}, optimized decode paths~\cite{hong2024flashdecodingpp}, and KV virtualization~\citep{kwon2023pagedattention}.

%% file: Content/4-evaluation.tex
\section{Evaluation}
\label{sec:evaluation}

We evaluate whether \MAC\ (i) preserves task quality relative to \emph{full attention}, (ii) delivers end‑to‑end efficiency gains \emph{in decode} with real serving frameworks, and (iii) derives its gains from its intended mechanisms (pre‑RoPE matching, L2 metric, local‑band rectification). Unless otherwise noted, \MAC\ is enabled for \emph{decode only}. \MAC\ also applies to prefill and is fully functioning, but may require additional tweaking, see results and analysis in \S\ref{app:prefill_decode}.

\subsection{Experimental Setup}
\label{subsec:setup}
\textbf{Models.} As shown in Table~\ref{tab:acc_8b}.

\textbf{Runtime.} SGLang~0.4.9 with FlashInfer~0.2.7 kernels, running on NVIDIA H100 SXM5 with CUDA~12.8.1.

Besides Quest, we also compare our method to other efficient attention strategies, e.g. SnapKV, TOVA, in Fig.~\ref{fig:compare}. For all experiments, we fix the random seed, set the temperature to 0, and keep only the attention path different.

\subsection{Benchmark details}
\textbf{LongBench v2}~\cite{bai2024longbenchv2} is a diverse long‑context suite (up to 120K tokens in our experiment) spanning QA, summarization, and retrieval tasks; results are partitioned into \emph{Short/Medium/Long} to evaluate length sensitivity.

\textbf{RULER}~\cite{hsieh2024ruler} provides controlled synthetic probes (fixed at 120K in our experiment) that stress long‑range retrieval, delayed recall, and robustness to positional biases and length extrapolation.

\textbf{LongGenBench}~\cite{wu2024longgenbench} targets \emph{continuous long‑form generation} with outputs up to 16K tokens, using an LLM as judge (default LLaMA-3.3-70B). It is particularly stringent for KV‑efficient methods: small perturbations in attention can accumulate over thousands of decode steps, producing cascading errors. To our knowledge, prior KV‑efficiency works have rarely reported evaluations on this benchmark; our study fills this gap.

Across benchmarks, we mostly use a fixed search window $K=1024$ with a rectification band $r=256$, and a threshold $\tau=0.45$ for context-heavy tasks and $\tau=0.75$ for generation-heavy tasks. More detailed on how to choose these parameters can be found in \S\ref{app:tradeoff-figure}, \S\ref{app:band_window_heatmap}, and \S\ref{app:practical}.

\subsection{Task Fidelity Relative to Full Attention}
\label{subsec:fidelity}
In Table~\ref{tab:acc_8b}, we report accuracy/finish rates on these three benchmarks. Under identical decode settings, \MAC\ ~\emph{matches} (and in some cases, slightly improves) full‑attention quality (we suspect it is due to some potential noise in long context computation for the baseline full attention), while achieving significant KV‑budget reductions. Gains concentrate in the Medium/Long buckets of LongBench v2 and carry over to LongGenBench with comparable finish rates; the few regressions (e.g., RULER at 70B) are small.


\noindent\textbf{Compared to selection.} Relative to token‑selection baselines (Quest), \MAC\ typically attains equal or higher accuracy at similar or stricter KV budgets—consistent with preserving access to all tokens rather than compressing/evicting them.

\subsection{Comparison on accuracy vs. speedup}
\label{app:more_comparison}
\begin{table}[t]
\centering
\caption{Overall accuracy on LongBench v2}
\resizebox{\columnwidth}{!}{%
{\fontfamily{ppl}\selectfont\footnotesize
\begin{tabular}{r r r r r r}
\toprule
\textbf{KV \%} & \textbf{Full Attn.} & \textbf{Quest} & \textbf{RocketKV} & \textbf{Multipole} & \textbf{MAC-Attn.} \\
\midrule
1  & 29.0 & 27.6 & 29.4 & 27.6 & \textbf{30.2} \\
5  & 29.0 & 27.8 & 29.2 & 27.8 & \textbf{30.4} \\
10 & 29.0 & 27.6 & 29.2 & \textbf{30.2} & \textbf{30.2} \\
20 & 29.0 & 28.2 & 29.4 & 28.0 & \textbf{29.6} \\
\bottomrule
\end{tabular}
}}
\label{tab:new_baseline_acc}
\end{table}

\begin{table}[t]
\centering
\caption{End-to-end Attention Latency ($\mu$s) on 120K context}
\resizebox{\columnwidth}{!}{%
{\fontfamily{ppl}\selectfont\footnotesize
\begin{tabular}{r r r r r r}
\toprule
\textbf{KV \%} & \textbf{Full Attn.} & \textbf{Quest} & \textbf{RocketKV} & \textbf{Multipole} & \textbf{MAC-Attn.} \\
\midrule
1  & 234.2 & 581.2 & 822.8  & 192.4 & \textbf{62.9} \\
5  & 234.2 & 594.7 & 844.7  & 210.8 & \textbf{64.0} \\
10 & 234.2 & 608.5 & 1042.5 & 265.4 & \textbf{78.1} \\
20 & 234.2 & 640.5 & 1855.6 & 324.6 & \textbf{103.8} \\
\bottomrule
\end{tabular}
}}
\label{tab:new_baseline_latency}
\end{table}

We evaluate more efficient attention methods on LongBench v2, and measure their end-to-end latencies at 120K context length. \textbf{Full Attn.} denotes the FlashInfer full attention baseline. For Quest~\cite{tang2024quest}, RocketKV~\cite{behnam2025rocketkv}, and Multipole~\cite{hooper2025multipole}, we followed the authors' official repositories to build kernels and used their official evaluation scripts for accuracy/latency. As shown in Table~\ref{tab:new_baseline_acc} and Table~\ref{tab:new_baseline_latency}, most existing efficient attention methods have significant overheads in addition to attention, thereby failing to surpass even the full attention FlashInfer baseline.

\subsection{Query Hit Rate Beyond Dense Models}
\label{subsec:moe-eval}

\begin{table}[h]
\centering
\caption{Evaluation on MoE model (Qwen3-30B-A3B-Instruct)}
\resizebox{\columnwidth}{!}{%
{\fontfamily{ppl}\selectfont\footnotesize
\begin{tabular}{lcccccc}
\toprule
\textbf{Model / Setting} & $\boldsymbol{\tau}$ & $\mathbf{K}$ & $\mathbf{r}$ & \textbf{Overall Acc.} & \textbf{Hit (\%)} & \textbf{Skip (\%)} \\
\midrule
Full attention & -- & -- & -- & 37.0 & -- & -- \\
\midrule
MAC-Attention & 0.45 & 512  & 256 & 37.0 & 99.5 & 98.9 \\
MAC-Attention & 0.45 & 1024 & 256 & 36.6 & 99.6 & 99.0 \\
MAC-Attention & 0.45 & 2048 & 256 & 37.6 & 99.6 & 98.8 \\
MAC-Attention & 0.45 & 4096 & 256 & 36.6 & 99.7 & 98.6 \\
\bottomrule
\end{tabular}
}}
\label{tab:moe-eval}
\end{table}

Mixture-of-Experts (MoE) models introduce dynamic expert routing, therefore, we evaluate our method on the MoE model \textbf{Qwen3-30B-A3B-Instruct} using the same threshold ($\tau = 0.45$). Table~\ref{tab:moe-eval} shows that MAC-Attention consistently achieves a $\geq 99\%$ hit ratio while maintaining comparable accuracy to full attention. This indicates that MoE expert routing does not significantly hinder effective semantic query matching in practice.

\paragraph{Discussion.}
Across all window sizes, MAC-Attention maintains stable accuracy relative to full attention while achieving very high hit and skip ratios. This suggests that, despite the conditional computation introduced by MoE routing, the semantic structure exploited by our matching mechanism remains sufficiently consistent for effective reuse.

\begin{figure}[t]
  \centering
  \includegraphics[width=\columnwidth]{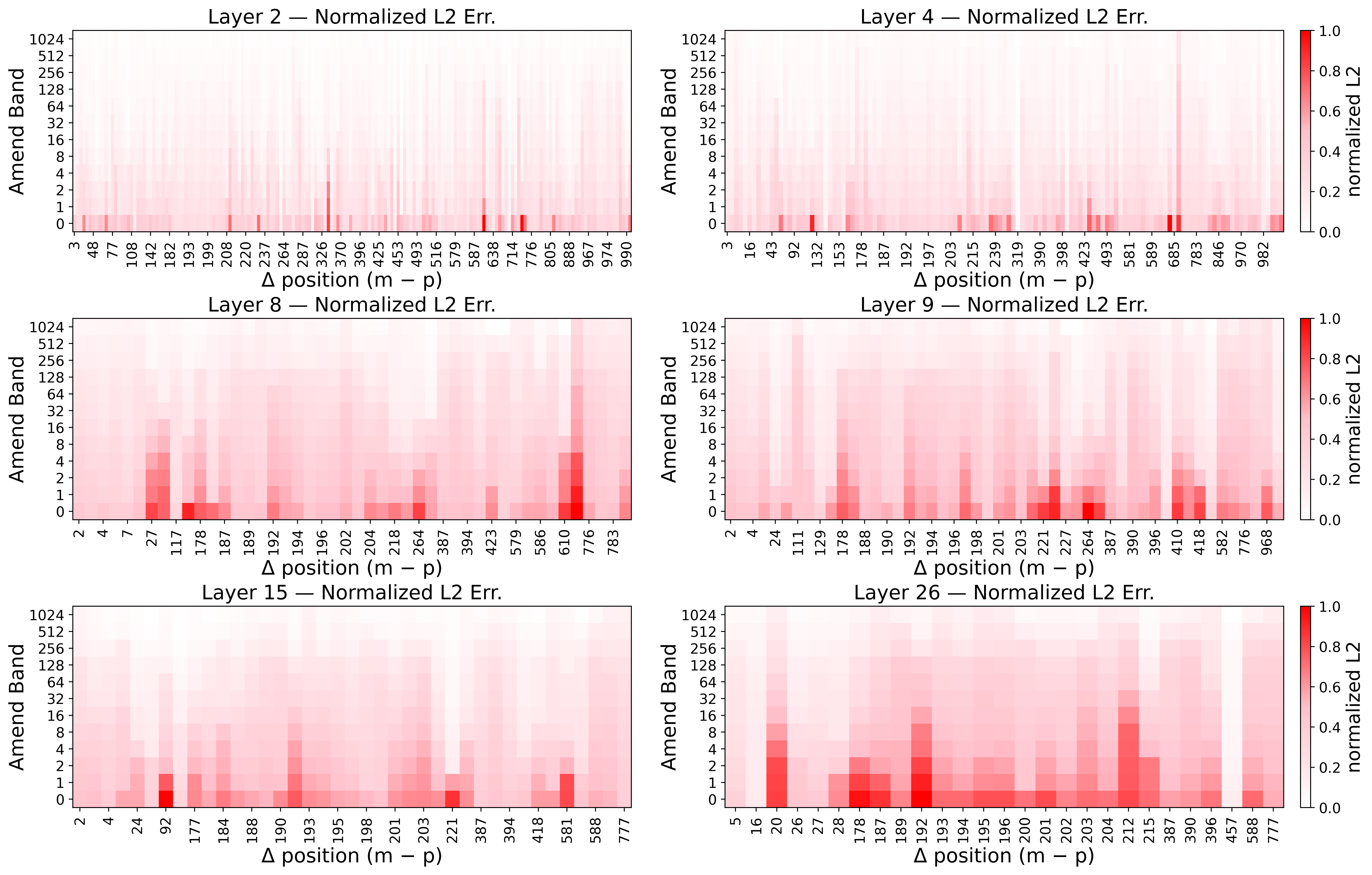}
    \caption{\textbf{Rectification error vs.\ reuse gap and band width.} Layerwise heatmaps show the normalized output error as a function of reuse gap (\(\Delta\)) and rectification band width (\(r\)); representative layers are displayed.}
  \label{fig:amend-err}
\end{figure}

\subsection{Rectification Error vs.\ Reuse Gap and Band Width}
\label{subsec:amend_error}

\paragraph{Setup and metric.}
We quantify how much \emph{amend} reduces approximation error as a function of the
rectification band width \(r\) and the reuse gap \(\Delta=m{-}p\), where
\(m\) is the current decode position and \(p\) is the matched query position
selected by the pre‑\RoPE\ L2 matcher from \S\ref{subsec:match}.
For each layer we compute the per‑token output error
\[
\mathrm{Err}(m)\;=\;
\frac{\big\|o^{\text{MAC}}_m-o^{\text{full}}_m\big\|_2}
     {\big\|o^{\text{full}}_m\big\|_2},
\]
and average over LongGenBench sequences and over all query heads
(\emph{GQA}: 32 Q heads sharing 8 KV heads).
Figure~\ref{fig:amend-err} visualizes the average error as a heatmap for
a few representative layers of LLaMA‑3.1‑8B (32 layers):
layers 2, 4, 8, 9, 15, and 20.
The x‑axis is the relative match position \(\Delta\), the y‑axis is the
rectification band width \(r\), and lighter color indicates lower error.

\textbf{Insights.} Error decays rapidly as the band widens, \(r\approx 8\) removes most discrepancy, and \(r\ge 256\) is effectively indistinguishable from full attention, indicating that recomputing a short high‑mass band captures the dominant discrepancy near the reuse boundary.
At fixed \(r\), larger reuse gaps are more sensitive, and deeper layers tend to require slightly wider bands—both consistent with position‑induced logit drift and layerwise aggregation.
Occasional outliers arise when residual attention mass extends beyond the amended band or when query differences behave like near‑uniform logit rescaling.
Overall, a small, fixed \(r\) per head suffices to make reuse faithful across layers and sequence lengths.

\subsection{Layerwise Reuse Patterns: Acceptance and Skipped‑Prefix Fraction}
\label{subsec:per_layer_variability}

\begin{figure}[t]
  \centering
  \includegraphics[width=\columnwidth]{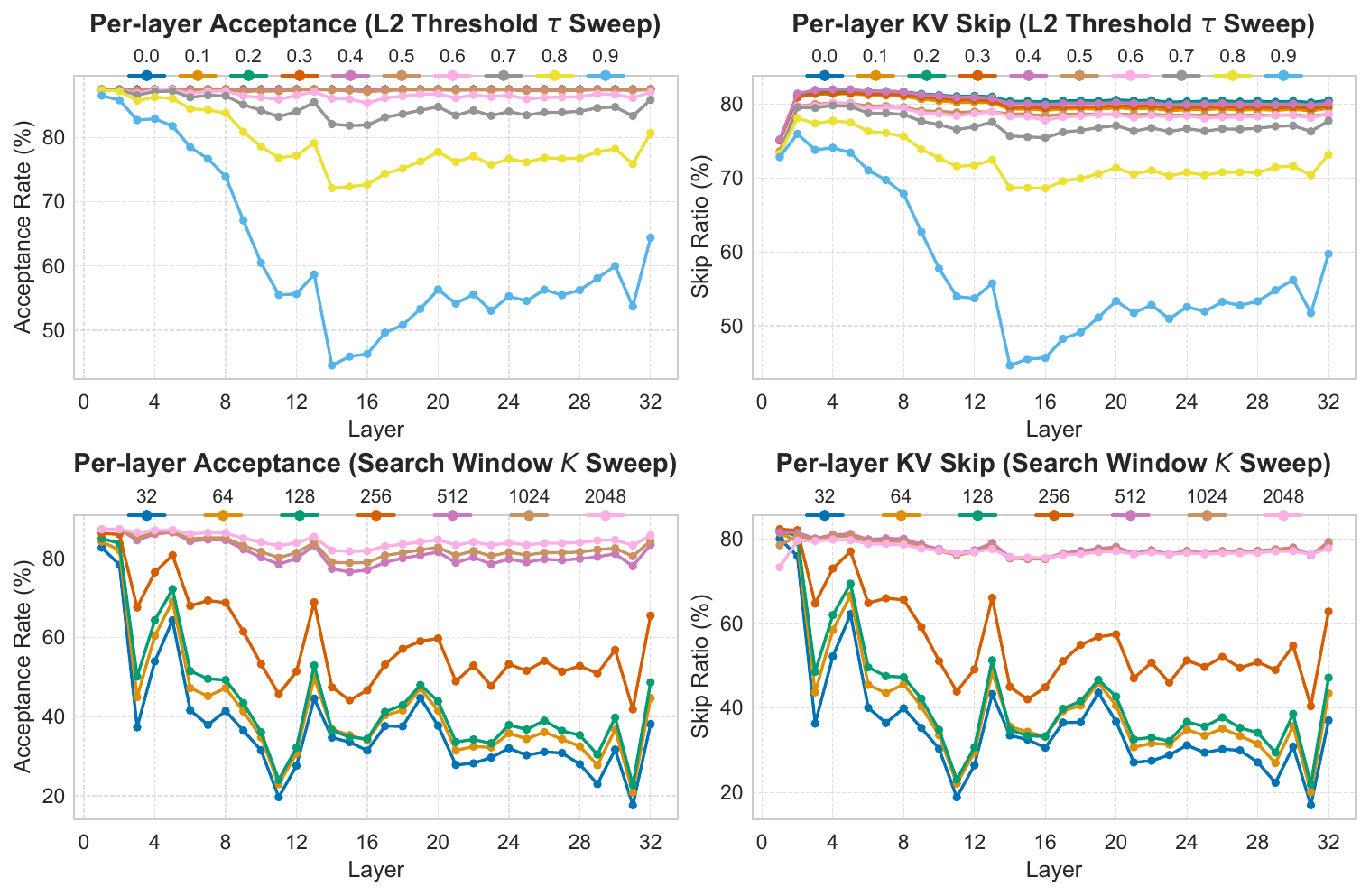}
    \caption{\textbf{Layerwise reuse patterns.} Acceptance rate (left) and skipped‑prefix fraction (right) per layer. Top: threshold sweep at fixed window size. Bottom: window‑size sweep at fixed threshold.}
  \label{fig:per_layer_acp}
  \vspace{-1em}
\end{figure}

\paragraph{Setup and metrics.}
We profile how often MAC can reuse a prefix \emph{per layer} using LongGenBench.
For a current decode position \(m\) and a matched index \(p\) (see \S\ref{subsec:match}), Figure~\ref{fig:per_layer_acp} reports:
(i) the \emph{acceptance rate}—the fraction of decoding steps for which at least one candidate in the cache window passes the pre‑\RoPE\ L2 test in Eq.~\eqref{eq:accept}; and
(ii) the \emph{skip ratio}—the expected fraction of the prefix that is not re‑read when a reuse occurs,
\(
\mathbb{E}\big[(p-r)_+/m\big]
\)
(we count misses as zero). All quantities are averaged over sequences and over the 32 query heads that share 8 KV heads (GQA) within each layer in the LLaMA-3.1-8B model.

\textbf{Threshold sweep (fixed window \(K{=}2048\))} varies the L2 threshold \(\tau\in[0,0.9]\) from Eq.~\eqref{eq:accept}, revealing layer dependent query similarity. \textbf{Window sweep (fixed threshold \(\tau{=}0.75\))} varies the search window \(K\in\{32,\ldots,2048\}\), revealing layer dependent temporal locality. 

\textbf{Insights.}
Reuse is strongly layer‑dependent.
Early layers exhibit high self‑similarity and thus frequent reuse; specific mid and upper layers are more variable and benefit from slightly larger windows.
Increasing the search window stabilizes acceptance and increases the skipped‑prefix fraction up to a point, beyond which returns diminish.
These profiles suggest that lightweight, per‑layer tuning of the match threshold and window size could further raise acceptance and skip ratios without meaningful overhead.

\begin{figure*}[thbp]
    \centering
    \includegraphics[width=0.98\textwidth,page=1]{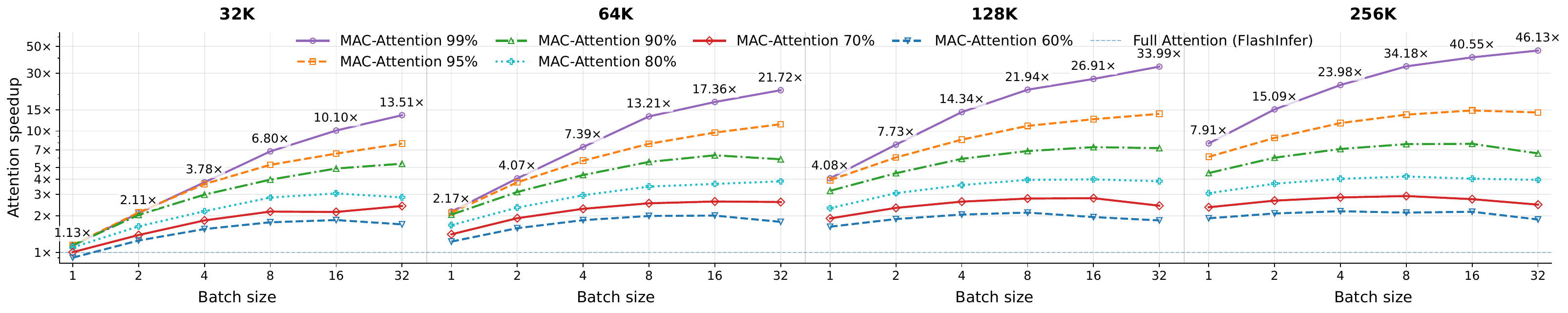}
    \caption{\textbf{Attention speedup vs.\ batch size and KV skip ratio across context lengths.}
    Speedup is computed as the \emph{combined attention‑phase latency} of our scheme (512-token matching window, load-balanced planning, rectified‑prefix reuse, and tail attention). The wide shared axes are split into four subgroups for contexts \{32K, 64K, 128K, 256K\}; within each subgroup, the x‑axis places batch sizes \(\{1,2,4,8,16,32\}\), the y‑axis (shown once, left) is log‑scale with a \(1\times\) reference line. Each line denotes a KV skip ratio. For more thorough latency comparison with existing efficient attention methods, refer to \S\ref{app:more_comparison}.}
    \label{fig:attn_speedup}
\end{figure*}

\begin{figure}[h]
  \centering
    \includegraphics[width=\linewidth]{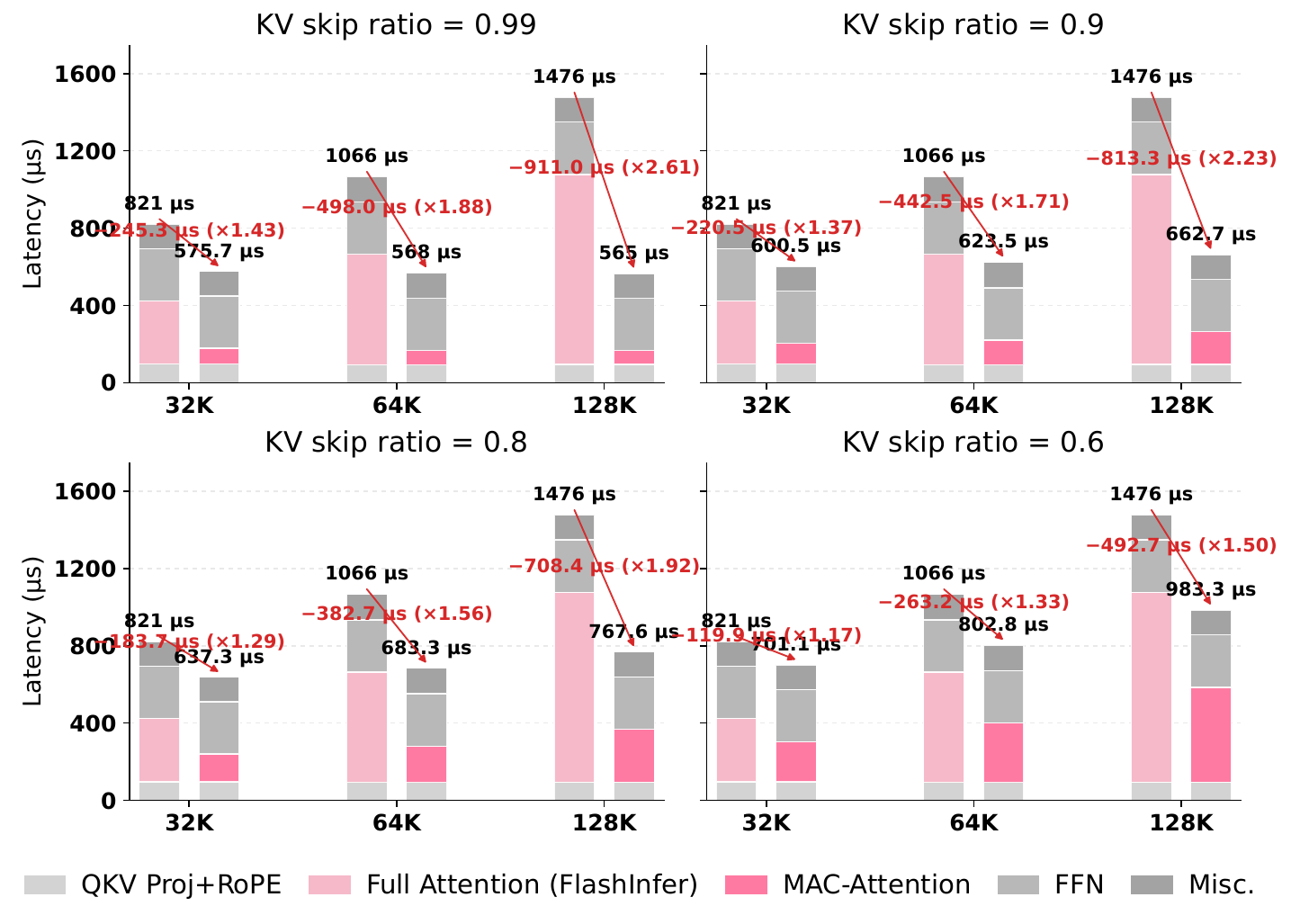}
    \caption{\textbf{End‑to‑end decode latency breakdown and speedup.}
For a fixed batch size, stacked bars show phase‑level latency (QKV+RoPE, attention, FFN, misc.) under full attention and \MAC\ across contexts and skip ratios; speedup is computed end‑to‑end.}
    \label{fig:b1-llama31-8b}
    \vspace{-1em}
\end{figure}

\begin{figure}[h]
  \centering
    \includegraphics[width=\linewidth]{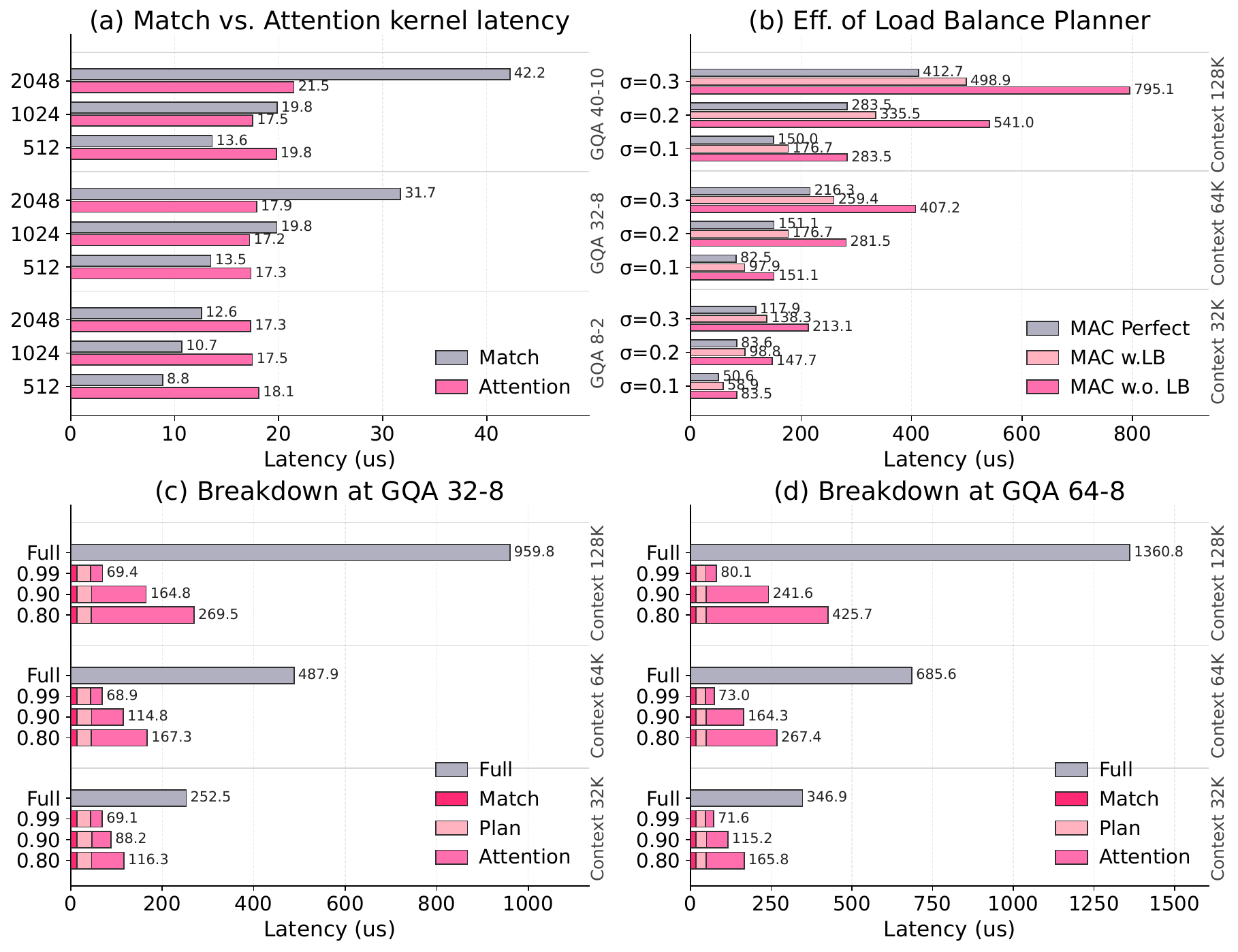}
    \caption{\textbf{Kernel‑level costs and load balancing.}
    \textbf{(a)} L2 \emph{Match} kernel latencies under different GQA settings and window lengths (using standard Attention for reference).
    \textbf{(b)} Effect of the load‑balancing planner compared with a perfectly balanced oracle and an unbalanced baseline.
    \textbf{(c–d)} Per‑step composition of \MAC\ (Match, Plan, Attention) across contexts.}
    \label{fig:kernel_breakdown}
    \vspace{-1em}
\end{figure}

\subsection{Fine-Grained Match/Attention Latency Profile}

\begin{table}[t]
\centering
\caption{Fine-grained attention-path latency profile ($\mu$s) at fixed match window \(K=1024\). Full attention reports FlashInfer attention latency. MAC reports the Match kernel, load balancer, attention kernel, and overall attention-path latency.}
\resizebox{\columnwidth}{!}{%
{\fontfamily{ppl}\selectfont\footnotesize
\begin{tabular}{c c c c c c c}
\toprule
\textbf{Context} & \textbf{KV \%} & \textbf{Full Attn.} & \multicolumn{4}{c}{\textbf{MAC-Attention}} \\
\cmidrule(lr){4-7}
 & & \textbf{Attention} & \textbf{Match} & \textbf{Load bal.} & \textbf{Attention} & \textbf{Overall} \\
\midrule
32K  & 1  & 71.6  & 9.1 & 28.8 & 26.0 & 63.9 \\
32K  & 5  & 71.6  & 9.1 & 28.9 & 25.2 & 63.2 \\
32K  & 10 & 71.6  & 9.1 & 28.6 & 25.1 & 62.8 \\
32K  & 20 & 71.6  & 9.1 & 29.5 & 27.3 & 65.9 \\
\midrule
60K  & 1  & 133.0 & 9.1 & 28.6 & 24.8 & 62.5 \\
60K  & 5  & 133.0 & 9.1 & 29.1 & 25.7 & 63.9 \\
60K  & 10 & 133.0 & 9.1 & 29.4 & 27.3 & 65.8 \\
60K  & 20 & 133.0 & 9.1 & 30.2 & 41.2 & 80.5 \\
\midrule
120K & 1  & 234.2 & 9.1 & 28.9 & 25.6 & 63.6 \\
120K & 5  & 234.2 & 9.1 & 30.1 & 25.5 & 64.7 \\
120K & 10 & 234.2 & 9.1 & 30.6 & 39.1 & 78.8 \\
120K & 20 & 234.2 & 9.1 & 30.2 & 65.2 & 104.5 \\
\bottomrule
\end{tabular}
}}
\label{tab:fine_grained_latency}
\end{table}

The match kernel runs only within the fixed search window \(K\), never over the full context. Thus its overhead is a constant regardless of the context length. Table~\ref{tab:fine_grained_latency} makes this explicit by showing the fine-grained latency profile of the Match kernel, load balancer, and attention kernel under 32K, 60K, and 120K contexts.

First, the Match kernel stays fixed at 9.1~$\mu$s across all contexts and KV-budget settings, confirming that it is bounded by the search window rather than the total sequence length. Second, the load balancer is also nearly constant, staying in the 28.6 to 30.6~$\mu$s range; this indicates that the scheduling overhead also does not scale materially with context length. Third, the MAC attention kernel is the only component that grows meaningfully, because it still depends on the effective band+tail span after reuse. At aggressive budgets (1\%--5\%), that kernel remains around 25~$\mu$s even at 120K, while at the more relaxed 20\% budget it grows from 27.3~$\mu$s at 32K to 65.2~$\mu$s at 120K.

Overall, these numbers show that MAC converts the dominant context-sensitive cost into a small constant front-end plus a reduced attention computation. For example, at 1\% KV budget the overall attention-path latency remains essentially flat, from 63.9~$\mu$s at 32K to 63.6~$\mu$s at 120K, whereas full attention increases from 71.6 to 234.2~$\mu$s. Even at 20\% KV budget, MAC rises only to 104.5~$\mu$s at 120K, still well below the full-attention baseline. This is the intended operating behavior: once the match window is fixed, the remaining latency growth is driven mainly by the unreused suffix rather than by the full prefix length.

\subsection{Attention‑Phase Speedup vs.\ Batch Size, Context Length, and Skip Ratio}

Figure~\ref{fig:attn_speedup} shows that \MAC’s gains grow systematically with (i) the KV skip ratio, (ii) batch size, and (iii) context length—exactly what we expect for an IO-bound workload. At fixed length and batch, higher skip (e.g., \textbf{MAC-99\%}) yields the largest acceleration because only $\sim1\%$ of the KV region is streamed and reduced per step; the remaining overheads—pre-RoPE matching, a small rectification band, and a constant-time log-domain merge—are largely length-independent. As batch size increases, the baseline saturates HBM bandwidth, so \emph{reducing KV bytes} translates into direct wall-clock gains for \MAC; for example at \textbf{32K}, the MAC-99\% series scales from $\sim1.1\times$ (batch 1) to $\sim13.5\times$ (batch 32). The effect compounds with longer contexts: at batch 32 and MAC-99\%, speedups climb from $\sim13.5\times$ (32K) to $\sim21\times$ (64K), $\sim34\times$ (128K), and $\sim46\times$ (256K).

At lower skip (e.g., \textbf{MAC-60\%}) and very small batches, the fixed reuse overhead can outweigh KV savings—hence a small regression vs.\ baseline at 32K/batch 1, suggesting a regression point where we should use full attention instead.

\subsection{End‑to‑End Decode Latency: Phase Breakdown and Speedup}
Figure~\ref{fig:b1-llama31-8b} shows that MAC’s end-to-end gains on LLaMA3.1-8B-Instruct increase monotonically with both context length and the KV skip ratio at a fixed batch size. \MAC\ only changes the attention path, with other operations untouched. This is the IO-bound regime we target: as sequences grow longer, attention (and thus KV traffic) occupies a larger fraction of the decode profile, so reducing KV bytes moved translates into proportionally larger wall-clock wins. Conversely, non-attention stages (QKV+RoPE, FFN, and runtime overheads) cap the overall speedup, consistent with Amdahl’s law. At high skip (e.g., $\gtrsim\!0.9$), MAC amortizes its fixed overheads (short-band rectification and constant-time log-domain merge), yielding substantially larger gains at 64K–128K than at 32K.

\subsection{Kernel‑Level Costs and Load Balancing Effects}

\paragraph{Kernel micro-benchmarks (Fig.~\ref{fig:kernel_breakdown}a).}
The L2 \emph{Match} kernel is inherently more memory-bound than FlashInfer \emph{Attention} under GQA because it scans per \emph{query} head while attention streams per (fewer) \emph{KV} heads. Consequently, with GQA \(8\!-\!2\) the matcher remains faster across lengths; at \(32\!-\!8\) it is faster at 512 but becomes slower at 1{,}024–2{,}048 as DRAM traffic dominates; and at \(40\!-\!10\) the trend amplifies, with match latency approaching \(\sim2\times\) attention at 2{,}048. Note, Figure~\ref{fig:kernel_breakdown}(a) is designed to evaluate the L2 \emph{Match} kernel and the reference \emph{Attention} kernel at the same token counts to expose their raw kernel behavior, since matching is more memory-bound under GQA. Across all experiments, however, the match window is fixed at \(K=1024\), so its latency is effectively constant regardless of the context length. Table~\ref{tab:fine_grained_latency} makes this explicit by showing the fine-grained latency profile of the Match kernel, load balancer, and attention kernel under 32K, 60K, and 120K contexts.

\paragraph{Load balancing efficacy (Fig.~\ref{fig:kernel_breakdown}b).}
Across contexts \(\{32\mathrm{K},\,64\mathrm{K},\,128\mathrm{K}\}\) and reuse skew \(\sigma\!\in\!\{0.1,0.2,0.3\}\), the CTA load-balancing planner consistently narrows the gap to an oracle ``perfect'' schedule. Relative to a naïve (unbalanced) assignment, load balancing trims attention-phase latency by \(\sim29\%\)–\(38\%\), and recovers \(\approx75\%\)–\(80\%\) of the \emph{excess} over the perfect baseline. A residual \(\sim16\%\)–\(21\%\) overhead remains due to CTA-granularity limits—once a CTA is partially filled, intra-CTA imbalance cannot be further corrected under the GQA scheme due to the workload mapping strategy in the current implementation. We conducted a more in-depth study to investigate the effect of this reuse skew in \S\ref{app:gqa_non_uniform}.

\paragraph{Component breakdown (Fig.~\ref{fig:kernel_breakdown}c–d).}
The per-step \emph{Match} cost is flat with respect to context length (bounded by fixed \(K\)); the \emph{Plan} stage is nearly constant and negligible. The \emph{Attention} component scales with length and dominates variability, decreasing predictably with higher KV skip. Hence end-to-end gains are driven by avoided KV bytes, growing with both skip ratio and sequence length, in line with the budget in Eq.~(\ref{eq:roi}).

\subsection{Performance fallback inside GQA KV group}
\label{app:gqa_non_uniform}
\MAC performs matching per query head, so different query heads within the same KV group may reuse different prefix points. Because our implementation maps a thread block to a KV group and loads KV once for all heads in the group, a single head with a low KV skip ratio directly increase work for that group.

\begin{table}[h]
\centering
\caption{MAC Attention matching skew within each KV group at different model layers. Numbers are rounded to integers.}
{\fontfamily{ppl}\selectfont\footnotesize
\begin{tabular}{c c c @{\hspace{1.5em}} c c c}
\toprule
\textbf{Layer} & \textbf{Avg.} & \textbf{Std.} &
\textbf{Layer} & \textbf{Avg.} & \textbf{Std.} \\
\midrule
0  & -399 & 0   & 18 & -278 & 17  \\
6  & -369 & 64  & 24 & -483 & 121 \\
12 & -272 & 15  & 31 & -346 & 40  \\
\bottomrule
\end{tabular}
}
\label{tab:match_skew}
\end{table}

The latency numbers of \MAC we reported already take into account this potential fallback. As shown in Fig~\ref{fig:kernel_breakdown}(b), the gap to the perfect balance is due to this non-uniformity. To quantify this, in Table~\ref{tab:match_skew} we measured within-KV-group match-position skew across layers on LongBench v2. The skew is typically small with some outliers. \textbf{Avg.} measures the averaged median of the relative position of successful matches, e.g., -399 denotes matching to a query that is 399 tokens prior to the current decoding token. \textbf{Std.} measures the median non-uniformity of these relative positions within each KV group. In the first layer, surprisingly, we observed a uniform matching, where all attention heads match to the same previous query. In other layers, we observed some fluctuations in matching positions.

\subsection{Auxiliary HBM Overhead of Query and Summary Caches}
\label{subsec:memory_overhead}

An important property of \MAC\ is that its extra state is bounded by the search window \(K\), rather than by the full context length \(L\). Besides the baseline KV cache, we only keep a ring of recent queries for matching and the associated cached summary bookkeeping within the same window. Therefore, the auxiliary HBM footprint grows as \(O(K)\), whereas the baseline KV cache grows as \(O(L)\). In practice, the query ring is the dominant term, while the cached summary scalars contribute only a small additional overhead.

\begin{table}[h]
\centering
\caption{Approximate auxiliary HBM overhead at 128K context.}
{\fontfamily{ppl}\selectfont\footnotesize
\begin{tabular}{c c}
\toprule
\textbf{Search window \(K\)} & \textbf{Aux.\ HBM overhead \%} \\
\midrule
256  & 1.2 \\
512  & 2.3 \\
1024 & 4.7 \\
2048 & 9.4 \\
\bottomrule
\end{tabular}
}
\label{tab:memory_overhead}
\end{table}

In our default setting \(K=1024\), the measured auxiliary footprint is only about \(5\%\) of the KV-cache size at 120K context, while the same setting delivers a \(7.9\times\) attention-phase speedup. This is the key systems tradeoff: \MAC\ pays a small, constant-size cache overhead in exchange for substantially reducing repeated KV traffic over the much longer active context.

Using the profiled \(K=1024\), \(L=120\mathrm{K}\) point as a reference, the relative auxiliary footprint can be approximated as
\[
\mathrm{Overhead}(K,L)\;\approx\;5\%\cdot\frac{K}{1024}\cdot\frac{120\mathrm{K}}{L},
\]
which reflects the linear dependence on \(K\) and the inverse dependence on the total context length \(L\). Table~\ref{tab:memory_overhead} reports the resulting auxiliary HBM overhead at \(L=128\mathrm{K}\). Even at \(K=2048\), the extra footprint remains below \(10\%\), which supports the practical operating regime used throughout the paper.

%% file: Content/5-related_works.tex
\section{Related Work}

\subsection*{Compression and selection}
KV compression cuts memory traffic via low‑rank/quantized caches and two‑stage pipelines~\cite{chang2024palu,zhang2024lorc,behnam2025rocketkv}, while selection/eviction prunes tokens or pages using statistics, heuristics, or learned signals—including position‑persistent sparsity~\cite{xiao2024streamingllm,zhang2023h2o,liu2023scissorhands,li2024snapkv,ge2023fastgen,feng2024adakv,tang2024quest,yang2025tidaldecode}; in contrast, \MAC\ keeps access to KV as a fallback or a correction in a very low frequency, avoiding prefix reads by reusing prior attention summaries, mathematically approaching the original full attention's fidelity.

\subsection*{Structural/prefix reuse and stepwise reuse}
Prefix reuse across requests and tree‑structured decoding shares computation when textual prefixes overlap~\cite{gim2024promptcache,yao2025deft}, and stepwise reuse alternates full and partial updates from prior attention patterns~\cite{xu2024recycled} which alternates full attention steps with partial steps that attend only to the top-$k$ tokens from a previously computed attention pattern; \MAC\ reuses within a single stream based on query similarity, independent of literal prefix overlap or decoding topology.


\subsection*{Systems and kernels}
IO‑aware kernels and serving systems improve locality, tiling, and paged‑KV memory for high‑throughput decode~\cite{dao2022flashattention,dao2023flashattention2,kwon2023pagedattention,hong2023flashdecodingpp,ye2025flashinfer}; \MAC\ composes with these stacks by reducing how much of the prefix must be used, without any training changes.

%% file: Content/6-discussion.tex
\section{Discussion and future Work}

In long context scenarios, we often observe that \MAC\ can achieve an acceptance/hit rate greater than 99\% within the $K$ search window, achieve a de-facto $O(1)$ complexity computation and memory access once hits. Although it still needs to visit all context on a match miss, falling back to $O(n)$ overall, we believe this scheme can be a promising direction for future research in efficient attention. More broadly, \MAC\ complements linear and sub‑quadratic attention by amortizing computation across time rather than altering representations or discarding tokens. As we suggested in \S\ref{subsec:amend_error} and \S\ref{subsec:per_layer_variability}, a more robust and adaptive configuration may further strengthen it.

%% file: Content/7-conclusion_and_future_work.tex
\section{Conclusion}

We presented a training‑free, model‑agnostic scheme, \MAC. It accelerates long context inference by reusing prior attention via a lightweight Match–Amend–Complete pipeline. The approach preserves full access to context, composes with IO‑aware kernels and paged‑KV managers. Across a wide range of long context benchmarks, \MAC\ maintains full‑attention quality while largely reducing KV reads, cutting end‑to‑end latency by over 60\% at 128K, and yielding up to $2.6\times$ faster token generation on LLaMA models (and up to $46\times$ on the attention phase). In practice, high match rates make step cost largely prefix‑insensitive, directly targeting the IO bottleneck that dominates long‑context decode. By reusing computation rather than compressing or discarding tokens, \MAC\ complements KV compression/selection and broadens the design space for fast, faithful attention.

%% file: Content/9-appendix.tex
\appendix
\section{Extended Experiments and Ablations}
\label{app:expts}
\label{app:impl}

This appendix is reorganized to present \emph{experimental results first}, followed by operator pseudocode and implementation analyses. We begin with
(i) the quality–efficiency trade-off curves (\S\ref{app:tradeoff-figure}),
(ii) the band–window heatmap for configuration selection (\S\ref{app:band_window_heatmap}),
and then (iii) an exploratory study enabling \MAC{} during both \emph{prefill} and \emph{decode} (\S\ref{app:prefill_decode}).
Subsequent sections provide the operator contract, numerics, data layout, pseudocode, tuning guidance, scheduling notes, diagnostics, and limitations.

\subsection{Quality–Efficiency trade-off curve}
\label{app:tradeoff-figure}

\begin{figure}[h]
  \centering
  \includegraphics[width=\columnwidth]{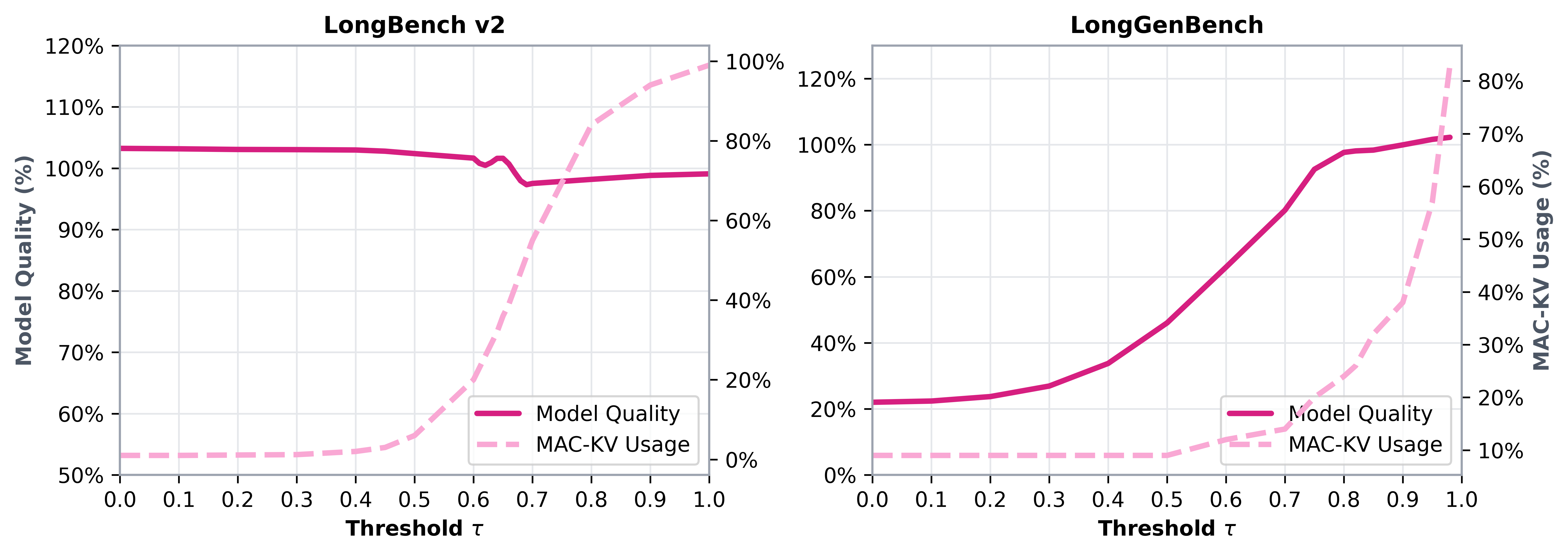}
  \caption{\textbf{Quality–efficiency trade-off vs.\ matching threshold \(\tau\).}
  Left: LongBench v2. Right: LongGenBench.
  Solid: quality (relative to baseline). Dashed: KV usage (lower is better).}
  \label{fig:model_quality_kv_usage}
\end{figure}

Figure~\ref{fig:model_quality_kv_usage} plots normalized model quality (solid) versus KV usage (dashed) as the matching threshold $\tau$ sweeps. The “knee” typically occurs where KV usage drops sharply while quality remains ${\approx}100\%$. Choose $\tau$ at or just before the knee, then adjust $K$ and $r$ minimally.

\subsection{Band–Window heatmap: normalized efficiency \& accuracy gating}
\label{app:band_window_heatmap}

\begin{figure}[h]
  \centering
  \includegraphics[width=\columnwidth]{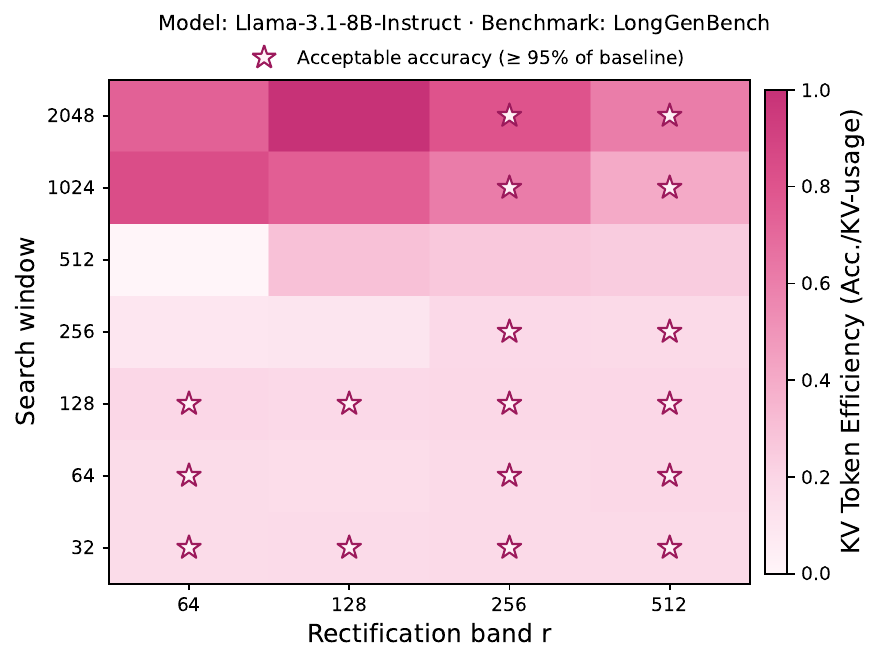}
  \caption{\textbf{Normalized KV token efficiency} \(\big(\text{Acc.}/\text{KV-usage fraction}\big)\) across \textbf{rectification band} and \textbf{search window}. Stars denote accuracy $\ge \theta$ of baseline.}
  \label{fig:heatmap_band_window}
\end{figure}

Figure~\ref{fig:heatmap_band_window} shows the normalized KV token efficiency
\(
\mathrm{Eff}_{\text{raw}}(r,K)=\mathrm{Acc}(r,K)/\mathrm{KVFrac}(r,K)
\)
over rectification band $r$ and window $K$. Colors are normalized within the grid (white$\to$pink: higher is better). A star marks configurations whose \emph{accuracy} satisfies
\(\mathrm{Acc}(r,K)\ge \theta\,\mathrm{Acc}_{\text{base}}\) (default $\theta=0.95$).

Empirically:
(i) very small windows ($K{\le}128$) look accurate but inefficient (few matches, little skipping);
(ii) larger windows expose longer reuse gaps and require larger bands (typically $r{\ge}256$) for fidelity;
and (iii) practical knees cluster near $(K,r)\!\in\!\{(1024,256),(1024,512),(2048,256)\}$.

\begin{table*}[t]
    \centering
    \caption{\textbf{Enabling \MAC\ for both prefilling and decoding on LongBench v2.}
    We sweep $\tau$ and band $r$; the \emph{window} is set to \texttt{inf.} to permit global search across the entire prefix (ablation-only; not production).
    Hit\% and Skip\% are reported as \texttt{prefill/decode}.
    Baseline is full attention.}
    {\fontfamily{ppl}\selectfont
        \begin{tabular}{ccc|cccccc}
        \toprule
        \multicolumn{3}{c}{Settings} & \multicolumn{6}{c}{LongBench v2} \\
        Threshold $\tau$  & Window & Band $r$ & Overall & Short & Medium & Long & Hit \% & Skip \%\\
        \midrule
        \multicolumn{3}{c}{Baseline (full attention)} & 29.0  & 35.0 & 26.0 & 25.0 & -- & -- \\
        \midrule
         0.30 & \texttt{inf.} & 512 & 26.6 & 30.6 & 25.1 & 23.1 & 99/99 & 55/86 \\
         0.35 & \texttt{inf.} & 512 & 26.8 & 34.4 & 22.8 & 22.2 & 99/99 & 55/86 \\
         0.45 & \texttt{inf.} & 128 & 25.6 & 29.4 & 21.4 & 27.8 & 96/95 & 56/85 \\
         0.45 & \texttt{inf.} & 256 & 27.2 & 33.3 & 22.3 & 26.9 & 96/95 & 55/85 \\
         0.45 & \texttt{inf.} & 512 & 27.2 & 30.0 & 21.9 & 33.3 & 96/95 & 54/84 \\
         0.55 & \texttt{inf.} & 512 & 26.8 & 30.6 & 21.4 & 31.5 & 86/83 & 49/74 \\
         0.60 & \texttt{inf.} & 512 & 27.4 & 31.7 & 23.7 & 27.8 & 77/71 & 45/62 \\
        \bottomrule
        \end{tabular}
    }
\label{tab:prefilling_decoding}
\end{table*}

\subsection{\MAC\ for both prefilling and decoding: exploratory study}
\label{app:prefill_decode}

\paragraph{Setting.}
Modern serving systems prefill in large chunks (e.g., 8K–16K tokens). We ablate enabling \MAC{} \emph{inside prefill} by letting each prefill token match against any earlier token (global window, denoted \texttt{inf.}), then amending a short band and completing with a merge—exactly as in decode. This explores feasibility; it is \emph{not} a production configuration.

\paragraph{Observations (Table~\ref{tab:prefilling_decoding}).}
Even with very high hit/skip rates, overall accuracy trails full attention when reuse spans very long gaps inside prefill. Length stratification shows that \MAC{} can help in the \emph{Long} bucket but may underperform in \emph{Short/Medium} buckets if the rectification band is too small. Increasing band size $r$ improves fidelity for long gaps but adds per-step cost.

\paragraph{Why prefill is harder.}
During prefill, the reuse gap $\Delta$ for late tokens in a chunk often spans the entire chunk because no within-chunk summaries exist yet. Post-position (post-\RoPE) logit drift accumulates across many positions; rectifying only a short band can leave residual error. By contrast, decode reuse typically happens over shorter horizons where a small fixed band suffices.

\paragraph{Recommendations.}
(i) \textbf{Distance cap}: accept a match only if the pre-\RoPE L2 test passes and $\Delta \le \Delta_{\max}$ (e.g., 1–2K). (ii) \textbf{Adaptive band}: scale $r$ with $\Delta$ or until the local mass outside the band is below $\varepsilon$ (e.g., $1-\rho \le 0.01$–$0.05$). (iii) \textbf{Two-gate acceptance}: combine pre-\RoPE L2 with a cheap post-\RoPE drift proxy (e.g., cached $\|Q_m-Q_p\|$). (iv) \textbf{Microprefill}: if supported, use smaller prefill chunks to reduce reuse gaps.

\paragraph{Bottom line.}
\MAC{} is robust for \emph{decode}. For \emph{prefill}, naive global matching with a short fixed band underperforms full attention overall despite high hit rates. Prefill viability improves with distance caps and adaptive bands.

\section{Practical Guidance, Scheduling, Diagnostics, and Assumptions}
\label{app:practical}

\subsection{Tuning guide}
\label{app:tuning}
\textbf{Window $K$.} Start with $K\in\{1024,2048\}$; increase only if mid/deep layers show low acceptance. Diminishing returns beyond ${\approx}4096$.

\textbf{Threshold $\tau$.} Under an isotropic null, pick $\tau$ to target a false-positive rate $\alpha$ via
\(
\alpha \approx F_{\chi^2_{d_q}}\!\big(d_q(1{-}\tau)^2\big),\;
\tau \approx 1-\sqrt{F^{-1}_{\chi^2_{d_q}}(\alpha)/d_q}.
\)
In practice, $\tau\in[0.4,0.8]$ works well globally; minor per-layer nudges can help.

\textbf{Band $r$.} Choose the smallest $r$ whose cumulative mass near the cursor exceeds $1{-}\varepsilon$ under baseline ($\varepsilon\in[0.01,0.05]$). Rules of thumb: $r\in[128,512]$; deeper layers may prefer slightly larger $r$.

\textbf{Dtypes.} Rings in \texttt{bf16}/\texttt{fp16} with \texttt{fp32} accumulators; keep $(m,Z)$ in \texttt{fp32}.

\subsection{Scheduling and concurrency}
\label{app:concurrency}
\textbf{Micro-pipeline.} Three kernels with minimal sync each step:
K1 L2-match (memory-bound, warp-reduced), K2 band+tail attention + merge (dominant), K3 rectify–append (aux stream).

\textbf{Auxiliary stream and graphs.} Launch K3 on an auxiliary CUDA stream with a single event wait; its outputs are consumed when the same layer is revisited. Capture steady-state into a CUDA Graph to amortize launches.

\textbf{Load balancing.} Heads have heterogeneous spans; flatten the (request, head) axis, assign CTAs proportional to span length, cap CTA size to preserve occupancy. This recovers most of the gap to an oracle perfectly balanced schedule.

\subsection{Diagnostics, failure Modes, and safeguards}
\label{app:diagnostics}
\textbf{Log (per layer/head).}
(i) acceptance rate; (ii) skipped-prefix fraction $\mathbb{E}[(p{-}r)_+/m]$ (misses as $0$);
(iii) baseline band mass $\sum_{t=p-r+1}^{p}\alpha^{(p)}_t$ (periodic probe);
(iv) norms $\|\tilde q\|_2$, $\|v\|_2$ percentiles.

\textbf{Common issues and mitigations.}
(1) Low acceptance in specific layers: increase $K$ or relax $\tau$ \emph{for those layers only}.
(2) Quality drift at very long gaps: enlarge $r$ slightly or cap reuse by $\Delta$.
(3) Numerical glitches in downdate: enforce common baseline $m$, and fall back to full attention if $Z^{\mathrm{rect}}\!\approx\!0$.
(4) Memory pressure: reduce $K$; keep $(m,Z)$ in \texttt{fp32}.

\textbf{Determinism.} With fixed seeds and decode settings, \MAC{} is deterministic given fixed match decisions; misses execute the full attention baseline.

\subsection{Deployment scope and assumptions}
\label{app:scope}
\textbf{Decode-only augmentation.} \MAC{} modifies \emph{decode} and leaves \emph{prefill} and weights unchanged unless explicitly enabled (ablation in \S\ref{app:prefill_decode}).

\textbf{Serving stack.} Assumes an IO-aware attention kernel (FlashAttention-style) and a paged-KV manager; composes with MQA/GQA (matching is per query head; attention streams shared KV heads).

\textbf{Per-request state.} Two rings of capacity $K$ (we only cache what we search): a \emph{query ring} with pre-\RoPE\ queries and an \emph{attention-summary ring} with rectified prefix summaries.

\subsection{Limitations and when to back off}
\label{app:limits}
\MAC{} relies on short-horizon semantic repetition and recency-biased attention. Heads with global, flat attention or idiosyncratic behaviors may see few matches and naturally fall back to baseline. For extremely large reuse gaps, a small fixed band can under-rectify; enlarge $r$ or skip reuse beyond a distance cap.